\documentclass[journal]{IEEEtran}

\ifCLASSINFOpdf

\else
 
\fi

\usepackage{array}
\usepackage{amsmath}
\usepackage{graphicx}
\usepackage{upgreek}
\usepackage{multirow}
\usepackage{color}
\usepackage{pifont}
\usepackage{url}
\usepackage{bm}
\usepackage{threeparttable}
\usepackage{float}
\usepackage[colorlinks,linkcolor=blue]{hyperref}
\begin{document}
 
\title{Grafting Transformer on Automatically Designed Convolutional Neural Network for Hyperspectral Image Classification}

\author{Xizhe~Xue*,~Haokui~Zhang*,~Bei~Fang,~Zongwen~Bai,~and~Ying~Li
\thanks{
* represents equal contribution. Manuscript received April 17, 2022; accepted May 26, 2022. Date of publication June 8, 2022; date of current version June 20, 2022 This work was supported by the National Natural Science Foundation of China (61871460, 62107027), Natural Science Foundation of Shaanxi Province (2020JM-556), China Postdoctoral Science Foundation (2021M692006).
X. Xue and Y. Li are with the School of Computer Science, National Engineering Laboratory for Integrated Aero-Space-Ground-Ocean Big Data Application Technology, Shaanxi Provincial Key Laboratory of Speech \& Image Information Processing, Northwestern Polytechnical University, Xi'an, China. (email: xuexizhe@mail.nwpu.edu.cn, lybyp@nwpu.edu.cn). H. Zhang is with Harbin Institute of Technology, Shenzhen. (e-mail: hkzhang1991@mail.nwpu.edu.cn). B. Fang is with the Key Laboratory of Modern Teaching Technology, Ministry of Education, Shaanxi Normal University, Xi’an, China. (e-mail: beifang@snnu.deu.cn). Z. Bai is with with the School of Physics and Electronic Information, Yan’an University, Yan’an, China. (e-mail: ydbzw@yau.edu.cn). (Corresponding author: Ying Li.)}

}

\markboth{IEEE TRANSACTIONS ON GEOSCIENCE AND REMOTE SENSING, VOL. 60, 2022}%
{Shell \MakeLowercase{\textit{et al.}}: Bare Demo of IEEEtran.cls for IEEE Journals}

\maketitle

\begin{abstract}

Hyperspectral image (HSI) classification has been a hot topic for decides, as hyperspectral images have rich spatial and spectral information and provide strong basis for distinguishing different land-cover objects. Benefiting from the development of deep learning technologies, deep learning based HSI classification methods have achieved promising performance. Recently, several neural architecture search (NAS) algorithms have been proposed for HSI classification, which further improve the accuracy of HSI classification to a new level. In this paper, NAS and Transformer are combined for handling HSI classification task for the first time. Compared with previous work, the proposed method has two main differences. First, we revisit the search spaces designed in previous HSI classification NAS methods and propose a novel hybrid search space, consisting of the space dominated cell and the spectrum dominated cell. Compared with search spaces proposed in previous works, the proposed hybrid search space is more aligned with the characteristic of HSI data, that is, HSIs have a relatively low spatial resolution and an extremely high spectral resolution. Second, to further improve the classification accuracy, we attempt to graft the emerging transformer module on the automatically designed convolutional neural network (CNN) to add global information to local region focused features learned by CNN. Experimental results on three public HSI datasets show that the proposed method achieves much better performance than comparison approaches, including manually designed network and NAS based HSI classification methods. Especially on the most recently captured dataset Houston University, overall accuracy is improved by nearly 6 percentage points. Code is available at: \href{https://github.com/Cecilia-xue/HyT-NAS}{https://github.com/Cecilia-xue/HyT-NAS}.

\end{abstract}


\section{Introduction}
Remote sensing observation plays an important role in earth observation and has many applications in agriculture and military~\cite{review1,review2}. Among various remote sensing observation technologies, hyperspectral image (HSI) classification is a fundamental but essential technique.
Captured by the amounts of hyperspectral remote sensing imagers, the HSIs of hundreds of bands contain much richer spectral information than ordinary remote sensing images, and the characteristic of containing both spatial and rich spectral information makes HSIs very useful for distinguishing ground-cover objects. Due to this, the HSI classification technology is widely applied in various scenes, e.g., mineral exploration~\cite{carrino2018hyperspectral}, plant stress detection ~\cite{behmann2014detection}, and environmental science~\cite{transon2018survey}, etc. However, in HSIs, feature vectors containing thousands of bands can be extracted from each spatial pixel location. Such high-dimensional features, on the one hand, help classify the ground objects, and on the other hand, increase difficulty in feature extraction. Therefore, it is worth exploring how to efficiently extract features from HSIs. During past decades, various features extractions have been applied and designed to extract robust features from HSIs~\cite{sun2020fast, zheng2017dimensionality, hang2018dimensionality}. Very recently, Luo \textit{et.al} proposed a multi-structure unified discriminative embedding (MUDE) method, which overcomes the drawbacks of previous graph-based methods that only consider the individual information of each sample~\cite{luo2021dimensionality}. In MUDE, the neighborhood, tangential, and statistical properties of each sample are introduced by using neighborhood structure graphs.  Duan~\textit{et.al} considered the manifold structure and multivariate relationship of samples from HSI in their proposed method geodesic-based sparse manifold hypergraph (GSMH)~\cite{duan2021semisupervised}. The non-linear similarity of the distribution of the sample on the manifold space is measured with the geodesic distance to build a manifold neighborhood for each sample. The final method achieves promising performance.

Since the year of 2012, deep learning has been developing rapidly and achieving remarkable results in various fields. Inspired by this, researchers have brought the deep learning methods in solving the problem of HSI classification, and gained impressive performance. In 2013, Lin \textit{et al.} utilized PCA to reduce the dimensionality of HSI from hundreds of spectral dimensions to dozens, then extract deep features from a neighborhood region via SAE. From 2014 to 2015, Chen \textit{et.al} introduced another spectral dimension channel based on the~\cite{lin2013spectral}. This additional channel directly takes the spectral features extracted from the pixel to be classified as input, and its output is integrated with the spatial spectrum channels to form a dual-channel structure \cite{chen2014deep, chen2015spectral}. In the same period, some other methods tried to apply 1D and 2D-CNN in the HSI classification. Specifically, 1D-CNNs are used to extract deep spectral features~\cite{2016Spectral, hu2015deep}, and 2D-CNN are employed to extract deep spatial features from HSI blocks that have been compressed along the spectral dimensions~\cite{makantasis2015deep, yue2015spectral}. After 2017, deep HSI classification methods primarily focused on extracting spatial-spectral features. Some work construct a dual-channel network structure to obtain spectral features and spatial features separately, and then merge them to form spatial-spectral features~\cite{zhang2017spectral}. Additionally, 3D-CNN is also a popular choice to  capture the spatial-spectral joint features directly~\cite{li2017spectral, chen2016deep}. Since 2017, various optimized 3D-CNN have been applied on the HSI classification task~\cite{zhong2018spectral, 2019Hyperspectral}, besides which some transfer learning methods have also been drawn into the classification of HSI images~\cite{2019Hyperspectral, yang2017learning}

The deep HSI classification methods give full play to the ability of extracting robust features independently. These deep HSI classification approaches show a significant advantage in classification performance compared to traditional HSI classification algorithms. However, these deep HSI classification approaches face a problem. Specifically, the network architectures in these methods are manually designed. For deep learning methods, designing an efficient network architecture is difficult, time-consuming, labor-intensive and requires a lot of verification experiments. This problem is even more serious in HSI classification. Because HSIs data are very different from each other in the number of bands, spectral range and spatial resolution, the suitable architectures are also different for different HSIs data. Therefore, it is usually necessary to design different network architectures for different HSIs data.

Moving beyond manually designed network architectures, Neural Architecture Search (NAS) techniques~\cite{liu2018darts} seek to automate this process and find not only good architectures, but also their associated weights for a given image classification task. NAS provides an ideal solution to liberate people from the heavy work of network architecture design. Chen \textit{et al.}~\cite{chen2019automatic} first introduced DARTS into the HSI classification task. This work compressed the spectral dimension of HSIs to tens of dimensions through a point wise convolution, and then directly used DARTS to search a 2D CNN that is suitable for specfic HSI dataset. Later, Zhang \textit{et al.}~\cite{zhang20213d} made an in-depth analysis of the structural characteristics of HSIs and proposed 3D-ANAS. In their work, a 3D asymmetric CNN is automatically designed under a pixel to pixel classification framework, which overcomes the problem of redundant operation existing in the previous classification framework and significantly improves the model inferring speed. 

In this work, further improvements have been made on 3D-ANAS from two aspects. 1) In 3D-ANAS, an asymmetric decomposition convolution is introduced in the search space, considering the difference between the spatial resolution and the spectral resolution of the HSI. However, this distinction between space and spectrum is only reflected on the operation level and is not free enough on the search space level. To be more specific, the entire search space consists of a sequence of blocks, each of which contains a number of operations. 3D-ANAS takes some asymmetric decomposition convolutions and other common convolutions as candidate operations, therefore 3D-ANAS can only separably process spatial and spectral information in operation level. It is difficult for such an approach to incorporate some classic hand-designed experience. For example, in the classical manually designed HSI classification network SSRN~\cite{zhong2017spectral}, the operation is completely separated into spectral processing and spatial processing. So in this article, we have constructed a new and more efficient search space, with more freedom to process the differences between spatial and spectral information. 2) The pure CNN structure is good at capturing local information but ignores global information, which has been proven to be very important for a lot of vision tasks. Inspired by this, we attempt to further improve the performance of automatically designed networks by integrating global information through grafting transformer modules. Before classification, we captured the relative relationship of pixels in different spatial positions and used this relationship to fine-tune the spatial-spectral features to achieve better classification accuracy. The main contributions of this work include the following three aspects:

\begin{enumerate}
    \item  By analyzing the characteristics of HSI, we propose a NAS algorithm to automatically design CNN for HSI. Specifically, we proposed a novel hybrid search space, which contains two types of cells, including space dominated cell and spectral dominated cell. The entire search space is built on these two cells and can be divided into inner and outer spaces. The inner space determines the topology structure in the cell, and the outer space decides whether the space dominated cell or the spectral dominated cell is selected on the specific layer.
    \item To further improve the classification accuracy, we attempt to graft the emerging transformer module on the automatically designed CNN to add global information to local region focused features learned by CNN. Benefiting from the pixel to pixel classification framework we adopted here, the transformer module can be seamlessly grafted to the end layer of CNN. Such a grafted structure takes advantage of the ability of a transformer to capture inner relationship of pixels, while avoiding the difficulties of training a complete transformer.
    \item  Experimental results on three typical HSI classification datasets, including Pavia Center, Pavia University and Houston University have validated that the proposed approach obviously improves the classification accuracy of auto designed HSI classification approaches.
\end{enumerate}

The rest of this paper is organized as follows. Section II reviews related work. Our approach is elaborated in Section III. Section IV provides algorithm implementation details, extensively evaluates and compares the proposed Hy-NAS and HyT-NAS approaches with state-of-the-art competitors. Finally, we conclude this work in Section V.

\section{Related Work}
\subsection{Hyperspectral Image Classification via CNNs}

Recent years have witnessed growing interests in using CNNs to deal with HSI classification problem. The development of HSI classification based on CNNs has mainly gone through three stages. 

From 2015 to early 2016, researchers focused primarily on HSI classification based on 1D-CNNs and 2D-CNNs. The methods based on 1D-CNNs generally employ 1D-CNNs to perform convolution along the spectral dimension to extract spectral features~\cite{mei2016integrating, 2016Spectral}. Beyond methods based on 1D-CNNs, a series of 2D-CNNs based HSI classification approaches are with good prospects. Intuitively, regions surrounding the pixel can provide additional visual information facilitating the classification. After compressing HSIs to low-dimension, 2D-CNNs based methods crop a neighborhood patch around the pixel to be classified. Then, this patch is fed to a 2D-CNN to extract the spatial-spectral features~\cite{makantasis2015deep, yue2015spectral}. Compared with the 1D-CNN based approaches, the methods based on 2D-CNNs achieve higher accuracy.  However, the classification results of methods that only use 2D-CNNs may not keep structural information very well. Their visual results are much smoother than those of the 1D-CNNs methods. The second development stage mainly focuses on combining 1D-CNN and 2D-CNN to perform the HSI classification. Taking the advantages of 1D-CNN and 2D-CNN, the dual-channel CNN structure can further improve the accuracy of HSI classification~\cite{zhang2017spectral, yang2017learning}. The third stage is the 3D-CNN stage. Inspired by the 3D structure of HSIs, 3D-CNNs have been gradually used in HSI classification approaches. Such methods directly construct 3D-CNNs to extract the spatial spectrum features. Compared with those of dual-channel CNNs, the structures of 3D-CNNs are always more simple, intuitive and powerful~\cite{li2017spectral,chen2016deep}. 

In recent years, optimizing the structures of 3D-CNNs for HSI classification has become mainstream. For example, the introduction of efficient residual structure~\cite{zhong2018spectral}, lightweight design and so on~\cite{2019Hyperspectral,meng2021lightweight,jia2020lightweight}.  Based on the classical residual structure, Zhong \textit{et al.}~\cite{zhong2018spectral} integrated the spectral residual and spatial residual modules, and then constructed a HSI classification model SSRN based on the two residual modules. Zhang \textit{et al.}~\cite{2019Hyperspectral} developed a lightweight 3D-CNN to optimize the model structure and proposed two transfer learning strategies (cross-sensor and cross-modality) to handle the problem of small sample~\cite{2019Hyperspectral}. Zhao \textit{et al.}~\cite{meng2021lightweight} proposed a lightweight spectral-spatial convolution HSI classification module (LS2CM) to reduce network parameters and computational complexity.

\subsection{ Neural network architecture search}
To overcome the heavy burden in manually designing network architecture, researchers turn their attention to NAS, which can automatically and efficiently discover the neural architectures that are suitable for certain tasks. Recent years have witnessed the success of NAS algorithms in plenty of general computer vision tasks, such as image classification~\cite{zoph2018learning}, object detection~\cite{NASFCOS} and semantic segmentation~\cite{liu2019auto}. So far,  the development of NAS always happened in three phases: architecture search based on evolutionary algorithm (EA) , architecture search based on reinforcement learning (RL) and architecture search based on gradient. RL based methods ~\cite{zoph2018learning,zhong2018practical} often contain a recurrent neural network (RNN) to perform as a meta-controller, generating potential architectures.  In the NAS methods enlightened by EA algorithms~\cite{real2019regularized, liu2017hierarchical, song2020efficient}, a series of randomly constructed models are evolved into a better architecture through EA. However, most RL methods and EA methods suffer from heavy computational cost and are less efficient in searching stage. The gradient-based NAS methods are proposed recently and can alleviate this problem to some extent. The first attempt DARTS is proposed in~\cite{liu2018darts}. Unlike the EA and RL-based method that train plenty of student networks, DARTS merely trains one super network in the searching phase, reducing training workload significantly. Getting inspiration from DARTS, Chen et.al.~\cite{chen2019automatic} proposed a 3D Auto-CNN for HSI classification. In the preprocessing stage, 
3D Auto-CNN heavily compresses the spectral dimension of raw HSIs through point wise convolution. The search space of 3D Auto-CNN are made up of 2D convolution operations in fact. 

Very recently, Zhang et.al.~\cite{zhang20213d} proposed 3D-ANAS, in which pixel-to-pixel classification framework and 3D hierarchical search space are jointly used. In conventional patch-to-pixel classification frameworks, all information in a cropped patch is used to
classify a single pixel. In a pixel-to-pixel framework, all pixels in a cropped patch are classified in one iteration. Adopting a pixel-to-pixel classification framework reduces repeat operations, speeding up inference efficiency significantly. In the 3D hierarchical search space, all operations are in 3D structure and the widths of networks can be adjusted adaptively in this work according to the characteristics of different HSIs. Benefiting from these two points, 3D-ANAS achieves promising performance. Unfortunately, 3D-ANAS still has two shortcomings:
\begin{enumerate}
    \item Previous work has indicated that learning the spectral and spatial representations separately is beneficial to extracting more discriminative features, such as SSRN. Although various asymmetric convolutions in the search space of 3D-ANAS allow the fine-tuning of the convolution kernel size and receptive field along spectral and spatial dimensions, this adjustment is limited inside a cell. Adjusting the proportions of spectral and spatial convolutions across the entire network is infeasible in this framework. 
    \item The pure convolutional structure mainly focuses on local neighborhood information, while ignoring the global relationship information among the whole input patch, which is often critical for high-level classification task.
\end{enumerate}

To overcome these two issues mentioned above, we propose a new NAS method for HSI classification. Specifically, to address the first issue, we design a hybrid search space that consists of two kinds of cells. One is space dominated cell and another is spectrum dominated cell. The hybrid search space has more flexible structures in selecting spatial or spectral convolution than the search space proposed in 3D-ANAS. Aiming to solve the second problem, a light transformer structure is grafted to the end of CNN, playing a similar role as a CRF to dig out the relationship between pixels.

\newcommand{\whiteding}[1]{\ding{\numexpr171+#1\relax}}

\begin{figure*}
	\begin{center}
		\includegraphics[height=0.36\linewidth,width=0.72\linewidth]{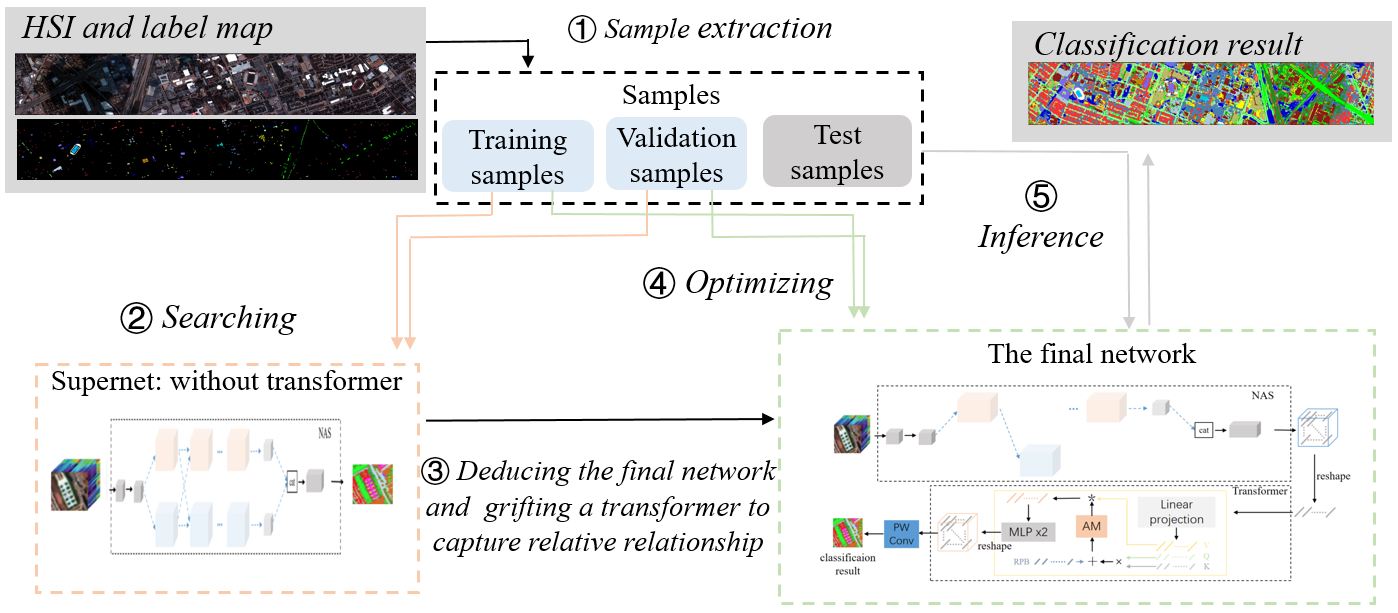}
	\end{center}
	\vspace{-0.3cm}
	\caption{ Workflow of proposed method. }
	\label{fig:fig1}
\end{figure*}

\subsection{Vision Transformer}
By in-depth analysis of the attention mechanism, Vaswani \textit{et al.}~\cite{vaswani2017attention} proposed the Transformer model. Compared with the RNN model previously applied to the NLP problem, Transformer improves the computational efficiency significantly. Because its structure can handle the elements in sequence in parallel. Besides, the Transformer inherits and further expands the ability of capturing the relationship between elements in the sequence, in comprehension with RNN. As a result, the introduction of Transformer has greatly promoted the development of NLP fields. 

In recent years, transformer models have been adopted in image processing and achieved very promising performance. Dosovitskiy \textit{et al.} proposed ViT~\cite{dosovitskiy2020image}, where the image is cut into patches then the patches are arranged into the input sequence for feature extraction. In order to keep sensitive to the position information of the patches, position embedding is introduced in the ViT. Besides, an additional class token is designed to perform the final classification. ViT's success in the fundamental visual tasks has greatly inspired the field of CV. Although the performance of ViT is relatively good, there still exist some problems, for instance, ViT has low computational efficiency and is hard to train. To alleviate the problem that the ViT is hard to train, Touvron \textit{et al.}~\cite{touvron2021training} proposed to use knowledge distillation to train ViT models, and achieved competitive accuracy with the less pre-training data. From the perspective of reducing computational cost and improving inference speed, Liu \textit{et al.}~\cite{liu2021swin} proposed the Swin transformer. Swin Transformer limits the calculation of attention to pixels within a small window,  which reduces the amount of calculation. Moreover, a shifted window based MSA is proposed, which makes the attention cross different windows. The Swin transformer has achieved higher accuracy than previous CNN models on tasks such as dense prediction. Very recently, after conducting a detailed analysis of the working principle of CNN and transformer, Graham \textit{et al.} mixed CNN and transformer in their LeVit model, which significantly outperforms previous CNNs and ViT models with respect to the speed/accuracy tradeoff~\cite{graham2021levit}.

Very recently, there are several methods adopting transformer models to classify HSIs~\cite{he2021spatial, qing2021improved, hong2021spectralformer}. In~\cite{he2021spatial}, He \emph{et al.}  designed a spatial-spectral Transformer, where a CNN is used to capture spatial information and a ViT is introduced to extract spectral relationship. Similarly, two parallel works also adopt Transformer to extract the spectral relationship. The network proposed in~\cite{qing2021improved} starts with a spectral relationship extraction Transformer and ends with several decoders.  In SpectralFormer~\cite{hong2021spectralformer}, a sequence of patches extracted from the input HSI is fed into Transformer.

Relevant to fusing the strength of CNN and the transformer model, our work is closely related to Levit. The difference is that the main body of our network still relies on an automatically designed CNN. In Levit, the transformer part is also the main part of feature extraction. The structure of the high-level CNN is equivalently replaced with the transformer structure. In our work, the transformer model is just to further capture the spatial relationship based on the features extracted by CNN. Compared to works that also adopt transformer models, our proposed HyT-NAS is a hybrid structure which inherits advantages of NAS and strengths of Transformer. Such structure makes it more stable and easier to train, while gaining better performance. For example, on the Houston University dataset, with only 450 training samples, our proposed HyT-NAS achieves 91.14\% overall accuracy, which is 3.13 percentage points higher than 88.01\%, the overall accuracy of SpectralFormer trained with 2823 training samples. In fact, such a phenomenon is consistent with the discovery presented in recent research works~\cite{graham2021levit, dai2021coatnet, zhang2022edgeformer, mehta2021mobilevit}. Hybrid structures generally achieve better performance than pure CNNs or ViTs as hybrid structures combine both advantages from CNN and ViT. In addition, previous Transformer structures adopted in HSI classification methods always focus on capturing spectral relationship, while in our work, the Transformer is responsible for capturing the relationship from all input space.


\begin{figure*}
	\begin{center}
		\includegraphics[width=5.6in ]{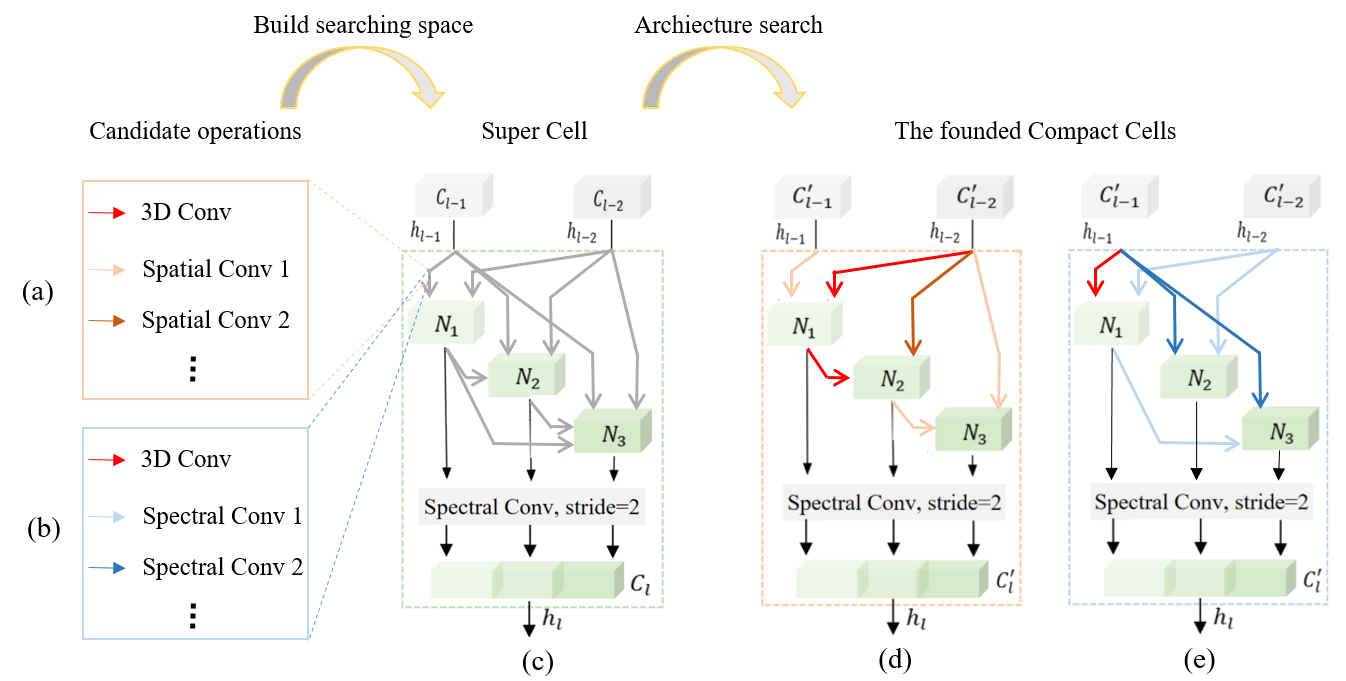}
	\end{center}
	\caption{ Architectures of supercells and the founded compact cells. (a) Candidate operations in space dominated supercell; (b) Candidate operations in spectrum dominated supercell; (d) Founded space dominated compact cell; (e) Founded spectrum dominated compact cell. In (a) and (b), different types of possible operation is represented as arrows with different colors. Topological architecture searching strategy aims to find a compact cell, in which each node only keep the two most valuable inputs, each of which is fed to the selected operation. For a more clear and concise presentation, we only show three nodes in the cell and three convolution operations in the search space. 
	}
	\label{fig:fig2}
\end{figure*}

\section{Proposed Method}

In this section, the proposed method is introduced in detail. First, as the proposed method contains more steps than previous deep learning based HSI classification approaches, we introduce the overall workflow briefly. Next, we elaborate on the proposed hybrid search space and compare it with the search space proposed in 3D-ANAS~\cite{zhang20213d}. Then we explain the reason for grafting the transformer module to the searched CNN and present the architecture of the grafted transformer module. Finally, we will briefly introduce our training process.

\subsection{Overall workflow}

As shown in Fig.\ref{fig:fig1}, the workflow of the overall classification framework can be divided into the following steps:

\noindent\textbf{1) Samples extraction:} Some pixels are randomly extracted from the whole HSIs according to certain proportions and rules. The collected sample pixels are divided into training set and validation set. The rest are reserved as test set. 

\noindent\textbf{2) Searching:} The collected training samples are fed into the CNN super network stacked by the space dominated cell and spectrum dominated cell. The training loss aims to minimize the loss between the prediction label and the groundtruth. The prediction accuracy of the network is validated on the validation set at a certain interval, and the loss and verification accuracy are recorded.

\noindent\textbf{3) Deducing the final network and grafting transformer:} The weight of the super network model with highest validation accuracy is used to deduce the final component network. According to the weight of the search model, the type of cell and the topology inside the cell are fixed in each layer. Besides, a flexible transformer structure is grafted at the end of the CNN network to capture the relationship between pixels.

\noindent\textbf{4) Optimizing the final compact network:} The training set is taken to optimize the grafted CNN-Transformer network structure, using the same loss as the searching stage.

\noindent\textbf{5) Inference:} After training, the compact model with the highest verification accuracy and the smallest loss is tested on the test set.

\subsection{Hybrid search space}

\noindent\textbf{Cell structure:} In 3D-ANAS, authors have already noticed that processing spatial and spectral information separately has better performance than using 3D convolution. In their work, classification accuracy is improved by introducing asymmetric search space. In this work, we further extend this discovery and propose hybrid search space, which consists of space dominated cells and spectrum dominated cells. As shown in Fig.\ref{fig:fig2}, The space dominated cell only contains spatial convolutions and 3D convolutions (instead of using the standard 3D convolution, we adopt separable 3D convolution as it has fewer parameters. In the following paragraphs, we call separable 3D convolution as 3D convolution for short), and the spectrum dominated cell includes some spectral convolutions and 3D convolutions. After searching, each layer can only keep one space cell or spectrum cell, and different layers do not share the cell structure. 

Compared with the search space proposed in 3D-ANAS, our designed hybrid search space is more flexible in selecting different operations to process spatial and spectral information. Note that this is relatively important for HSI datasets, as HSI datasets have a special characteristic, that is HSI datasets have different relatively low spatial resolution and extremely high spectral resolution. Roughly processing spatial and spectral information usually generates inferior classification accuracy.

\begin{figure*}
	\begin{center}
		\includegraphics[width=6.4 in]{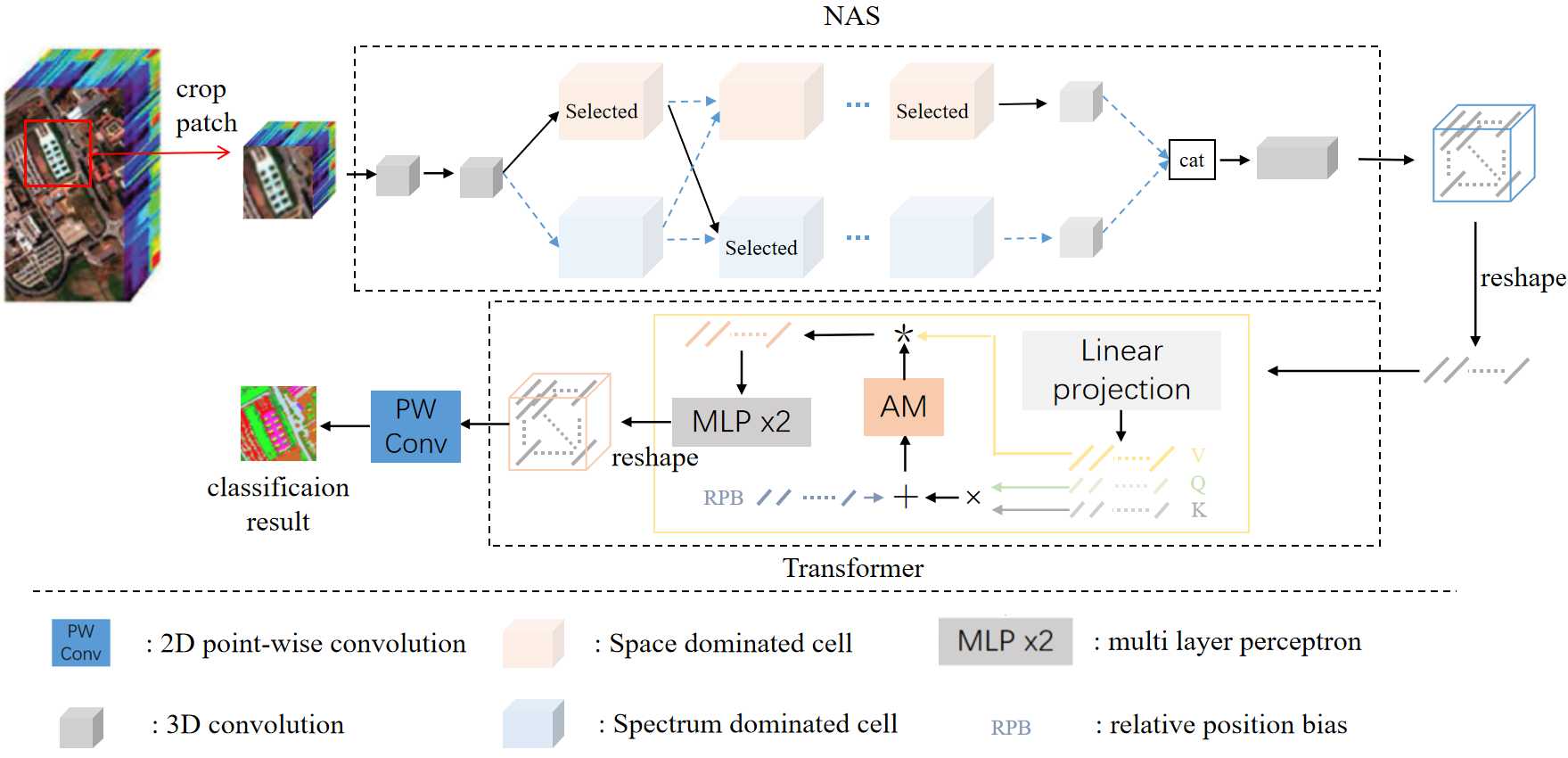}
	\end{center}
	\caption{Architecture of our final classification framework. Up: The searched compact network. The final selected blocks are marked as $Selected$; Down: The structure of grafted transformer. }
	\label{fig:fig3}
\end{figure*}

In specific, the space dominated cell includes the following operations:
\begin{itemize}
    \item acon\_3-1: LReLU-$\rm Conv(1\times3\times3)$-BN;
    \item acon\_5-1: LReLU-$\rm Conv(1\times5\times5)$-BN;
    \item asep\_3-1: LReLU-$\rm Sep(1\times3\times3)$-BN;
    \item asep\_5-1: LReLU-$\rm Sep(1\times5\times5)$-BN;
    \item con\_3-3: LReLU-$\rm Conv(1\times3\times3)$-$\rm Conv(3\times1\times1)$-BN;
    \item con\_3-5: LReLU-$\rm Conv(1\times3\times3)$-$\rm Conv(5\times1\times1)$-BN;
    \item skip\_connection: $f(x)=x$;
    \item discarding: $f(x)=0$.
\end{itemize}
The spectrum dominated cell includes the following operations:
\begin{itemize}
    \item econ\_3-1: LReLU-$\rm Conv(3\times1\times1)$-BN;
    \item econ\_5-1: LReLU-$\rm Conv(3\times1\times1)$-BN;
    \item esep\_3-1: LReLU-$\rm Sep(3\times1\times1)$-BN;
    \item esep\_5-1: LReLU-$\rm Sep(5\times1\times1)$-BN;
    \item con\_3-3: LReLU-$\rm Conv(1\times3\times3)$-$\rm Conv(3\times1\times1)$-BN;
    \item con\_3-5: LReLU-$\rm Conv(1\times3\times3)$-$\rm Conv(5\times1\times1)$-BN;
    \item skip\_connection: $f(x)=x$;
    \item discarding: $f(x)=0$.
\end{itemize}
where LReLU, BN, Conv and Sep represent LeakyReLU activation function, batch normalization, common convolution and separable convolution. 

\noindent\textbf{Architecture searching strategy:}
The network architecture search process can be divided into inner and outer search parts. The outer search part determines the cell type of this layer and the inner search strategy decides the cell's internal topology structure. The finally searched $L$-layer network may contain $L$ different cell structures and every cell contains a sequence of $N$ nodes. 

The inputs for each node consist of the outputs of all previous nodes and two inputs of the current cell. 
Assuming that each path in a cell contains all the $P$ candidate operations, the output of node $x_{i}$ is: 
\begin{equation}
x_{i}=\sum_{j=1}^{j=P}\left(\omega_{i}^{j} \bullet o_{i}^{j}\right)
\end{equation}
where $o$ and $\omega$ represent the different convolution operations and their corresponding weights, respectively. This weight is learnt through inner search according to the back propagation gradient. The output of a cell $h_{l}^{k}$ is obtained by:
\begin{equation}
\small
h_{l}^{k}=\operatorname{concat}\left( x_{i}^{k} \mid i \in\{1,2, \cdots, N\}\right).
\end{equation}
where $k$ denotes the cell type and $l$ represents the layer number. 

When optimizing the internal topology of a cell, the outer selection on cell types is also ongoing. Specifically, two types of cells are provided for each layer, focusing on spatial information and spectral information, respectively. In each layer, the outputs corresponding to two types of cells are weighted via learnable weights $\alpha_{i}$ and $\beta_{i}$, and then combined to form the cell output $h_{l}^{k}$. The output of layer $l$ can be expressed as:  
\begin{equation}
h_{l}=\operatorname{concat}\left(\alpha_{l}\cdot h_{l}^{spa}+\beta_{l}\cdot h_{l}^{spe}\right)
\end{equation}

After the stage of searching for network architectures, we build a compact network according to the learned structure parameters $\omega$, $\alpha$ and $\beta$. Specifically, for inner topology structure, we keep the two operations corresponding to the top two weights and prune the rest in each cell. For outer structure, we compare the $\alpha$ and $\beta$, then reserve the cell whose weight is bigger.

\subsection{The structure of Transformer}



So far, the proposed architecture has been a pure convolution structure. The final compact network founded by Hy-NAS is also a pure CNN. The Hy-NAS algorithm not only improves the external structure but also reserves the inductive bias of convolution operations. In other words, the final compact network founded by Hy-NAS also inherits the disadvantage of pure CNN. Pure CNN is good at extracting local features and ignores global relationships. Adding global information to local features always leads to much better performance, which has been verified by previous non-local related works~\cite{wang2018non} and the emerging transformer models~\cite{touvron2021training}. Especially for some dense prediction tasks, using global information may bring a significant improvement~\cite{wang2021pyramid}.

\begin{figure}
	\begin{center}
		\includegraphics[width=0.9\linewidth]{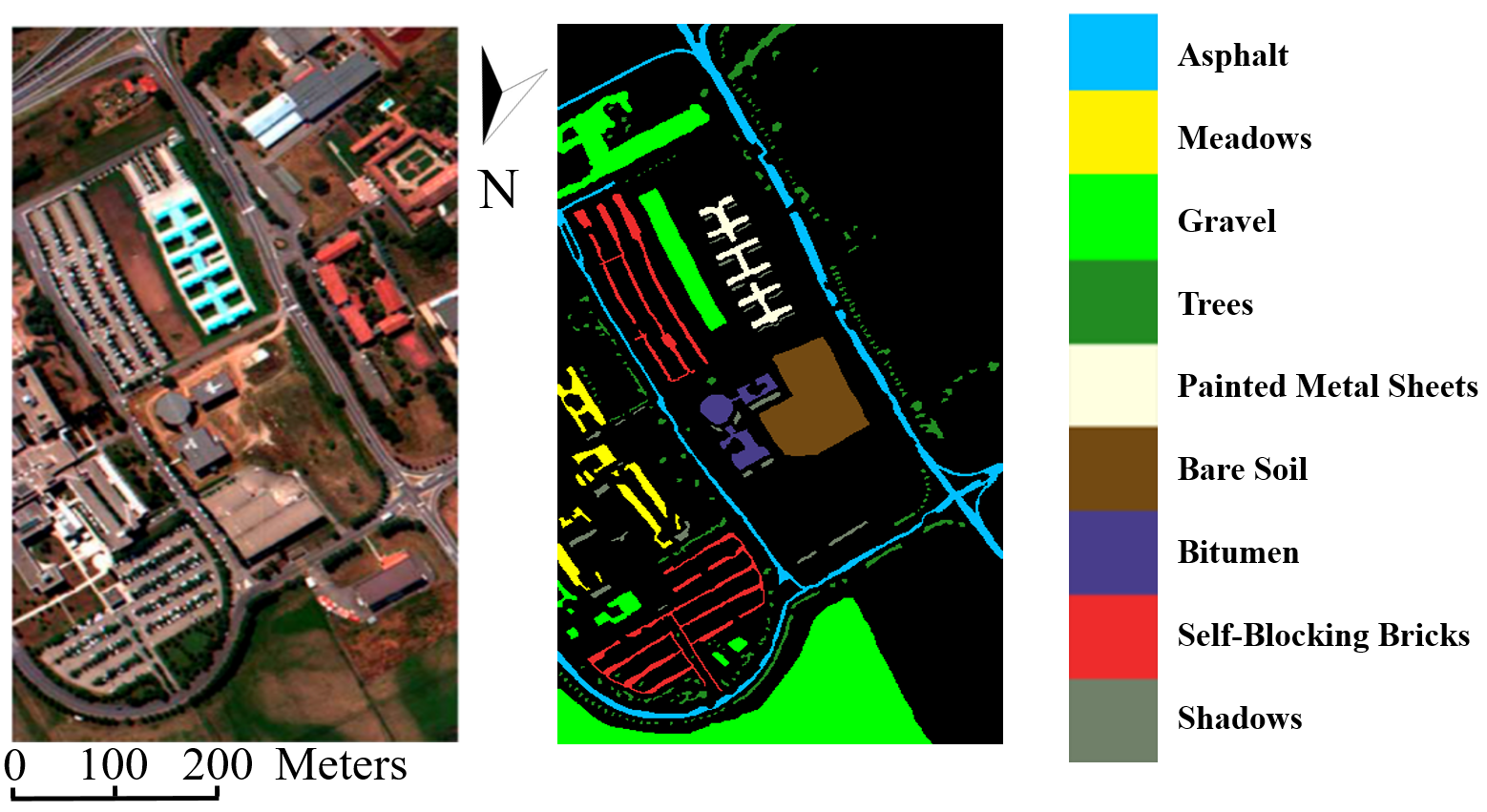}
	\end{center}
	\caption{ False color composites and ground truth maps of Pavia University. }
	\label{fig:PU}
\end{figure}

Therefore, to further improve the performance of Hy-NAS, we attempt to integrate global information to features learned by pure CNN. A natural idea is to add some non-local modules into the search space. As Hy-NAS is a NAS algorithm, adding non-local modules into search space does integrate information, but it also increases the complexity of search space and the difficulty of training super net. Here, we make a trade-off. Specifically, we graft a flexible transformer module at the end of the final compact network founded by Hy-NAS. Finally, we obtain a new HSI classification method, HyT-NAS. 

Such a grafting operation integrates global information to features learned by CNN, while avoiding introducing a complex search space which may improve the workload of architecture search. Inspired by the promising performance of transformer models, we choose to graft a transformer module to integrate global information. As shown in the bottom half of Fig.\ref{fig:fig3}. 

Before being split to sequence and sent to the transformer unit, the feature map $f$ of size$(B,C,W,H)$ from the encoder is reshaped and transposed to $f\in(B,N,C)$. The $Q,K,V$ is calculated through a linear layer and batch normalization layer (refers to linear projection in Figure~\ref{fig:fig3}). Here, linear-projection is responsible for mapping input vectors to three different feature spaces $Q$, $K$ and $V$, which play different roles in the following computational procedure. The definitions and functions of $Q,K,V$  can be found in~\cite{vaswani2017attention}. The $f$ is the input of the attention layer, and $f_{attn}$ is computed according to Eq.4:
 \begin{equation}
 {f_{attn}}=\operatorname{softmax}\left(\frac{Q K^{T}}{\sqrt{d_{k}}}+P\right) V
\end{equation}
where $d_k$ denotes the dimensionality of $V$, and $P$ means relative position embedding (RPB).
\begin{equation}
P_{(x, y),\left(x^{\prime}, y^{\prime}\right)}^{h}=Q_{(x, y),:} \cdot K_{\left(x^{\prime}, y^{\prime}\right),:}+B_{\left|x-x^{\prime}\right|,\left|y-y^{\prime}\right|}^{h} \cdot
\end{equation}
where $B^{h}$ represents the translation-invariant attention bias. The output $f_{\text {out }}$ of transformer can be generated as Eq.6 and then reshaped to the same dimensionality as the input $f$: 
\begin{equation}
f_{\text {out }}=\operatorname{MLP}\left(\operatorname{MLP}\left(A F\left(B N\left(f_{\text {att }}\right)\right)+f\right)\right)
\end{equation}
in which the $\operatorname{MLP}$ and $BN$ denote multi-layer perception and batch normalization layer, respectively. $AF$ means activation function. Specifically, a Hardswish function is employed in this work.

\begin{figure}[!t]
	\begin{center}
		\includegraphics[width=0.9\linewidth]{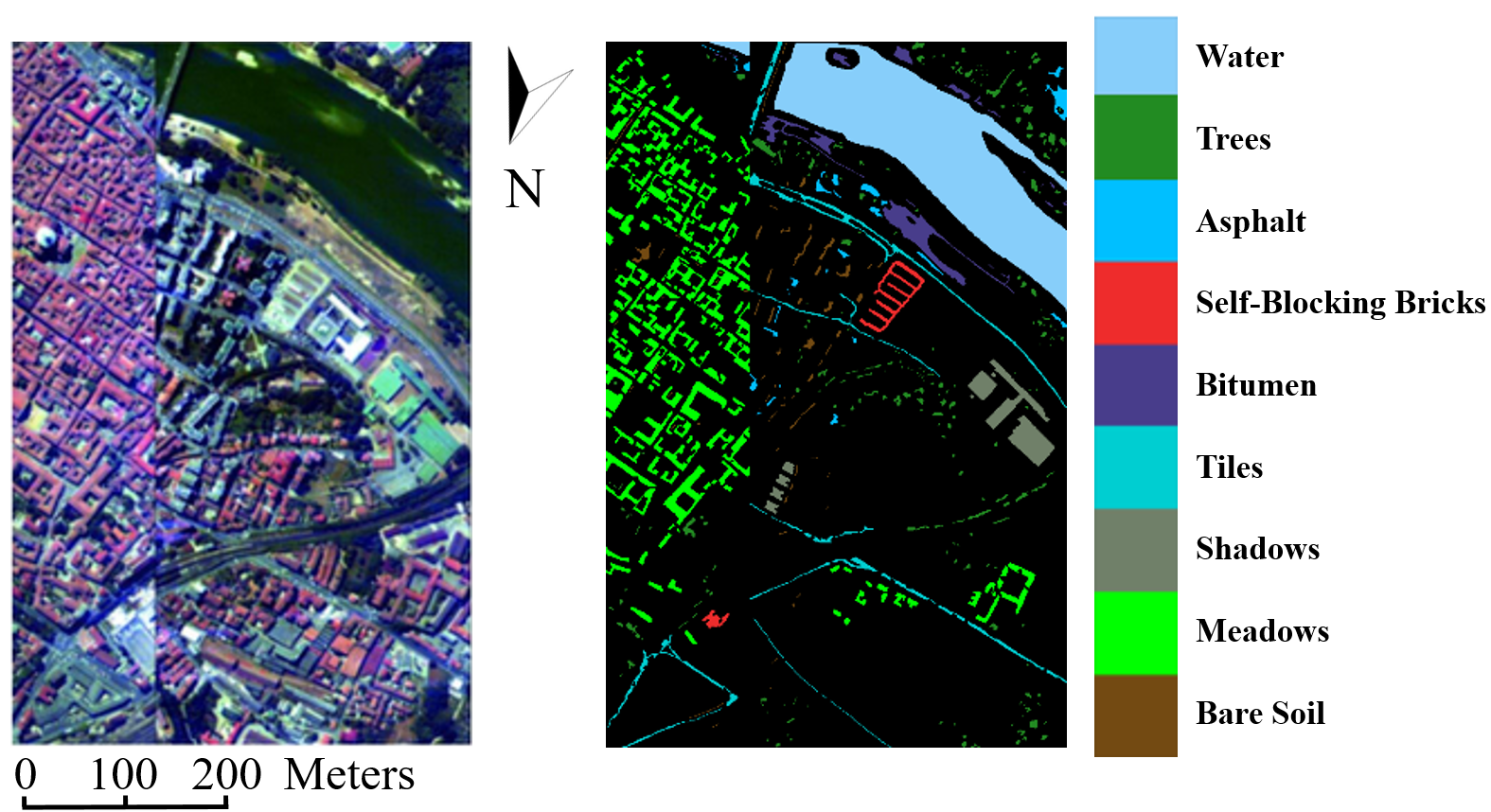}
	\end{center}
	\caption{False color composites and ground truth maps of Pavia Center. }
	\label{fig:PC}
\end{figure}

\subsection{Training Process}
In this work, we have followed the pixel-to-pixel classification framework of 3D-ANAS. Therefore, to fairly verify the effectiveness of the proposed contributions, we apply the same sampling rules, searching and training strategies as those in 3D-ANAS. After taking a 3D image cube from raw HSI and predicting the class of each 2D position in the cube, the cross entropy loss has been calculated according to the sparse training label map.
\section{Experiments}
Experiments are conducted on a server with an Intel(R) Xeon(R) Gold 6230 CPU @ 2.10GHz, 512 GB of memory, and Nvidia Tesla V100 32 GB graphics card. The training and testing experiments were implemented by using the open-source framework Pytorch 1.8  \footnote{ \url{https://pytorch.org/docs/1.8.0/}}

\begin{figure*}[htbp]
	\begin{center}
		\includegraphics[width=0.9\linewidth]{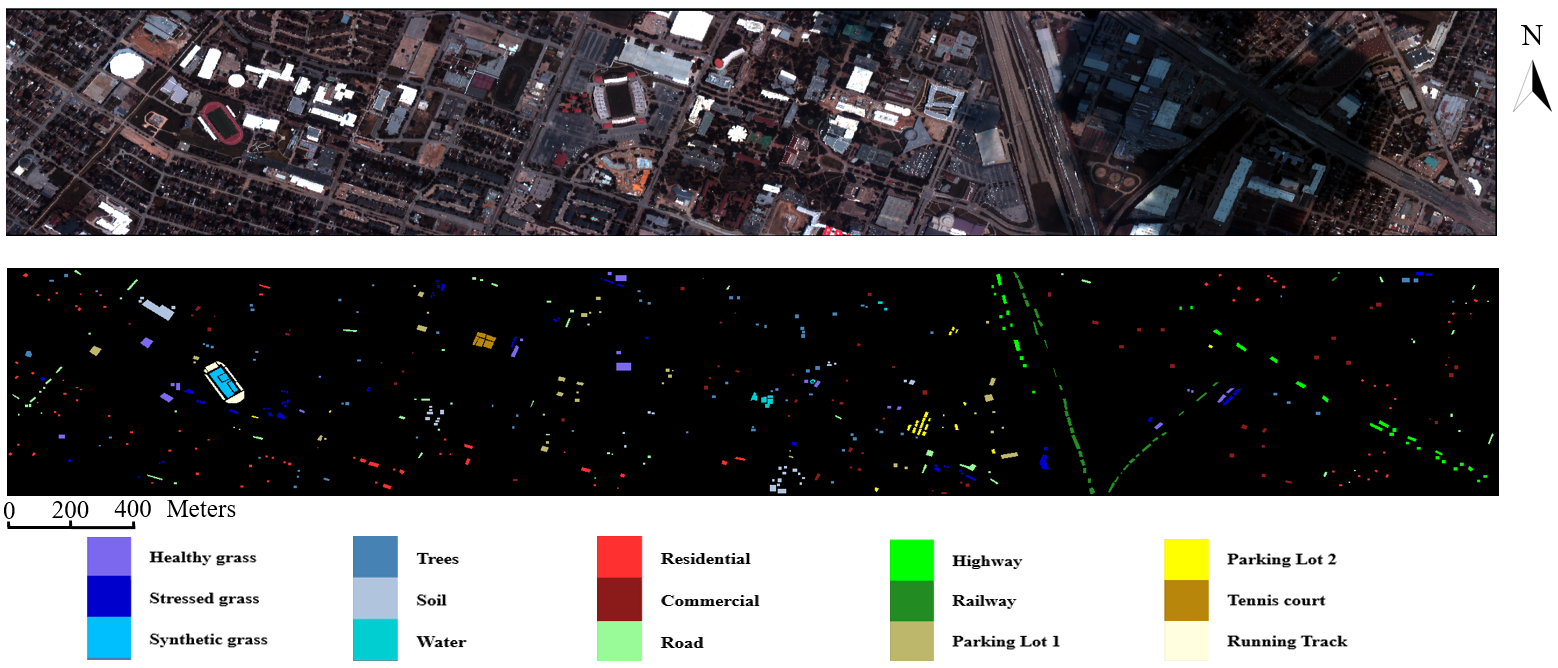}
	\end{center}
	\caption{ False color composites and ground truth maps of Houston University. }
	\label{fig:HU}
\end{figure*}

\begin{table*}[!t]
\setlength{\abovecaptionskip}{0.cm}
\setlength{\belowcaptionskip}{-0.cm}
\renewcommand{\arraystretch}{1.1}
\caption{Sample distribution information of Datasets}
\label{tab: HU}
\centering
\begin{tabular}{c|cc|cc|cc}
\hline
\multicolumn{3}{c}{Pavia University} & \multicolumn{2}{|c}{Pavia Center} & \multicolumn{2}{|c}{Houston University} \\
\hline
Class & Land Cover Type & No.of Samples & Land Cover Type & No.of Samples & Land Cover Type & No.of Samples \\
\hline
1	& Asphalt              &  6631  & Water                & 824   & Healthy Grass   &  1251 \\  
2	& Meadows              &  18649 & Trees                & 820   & Stressed Grass  &  1254 \\
3	& Gravel               &  2099  & Asphalt              & 816   & Synthetic Grass &  697  \\
4	& Trees                &  3064  & Self-Blocking Bricks & 808   & Trees           &  1244 \\
5	& Painted Metal Sheets &  1345  & Bitumen              & 808   & Soil            &  1242 \\
6	& Bare Soil            &  5029  & Tiles                & 1260  & Water           &  325  \\
7	& Bitumen	           &  1330  & Shadows              & 476   & Residential     &  1268 \\
8   & Self-Blocking Bricks &  3682  & Meadows              & 824   & Commercial      &  1244 \\
9	& Shadows              &  947   & Bare Soil            & 820   & Road            &  1252 \\
10  &  -                   &   -    &  -                   &   -   & Highway         &  1227 \\
11  &  -                   &   -    &  -                   &   -   & Railway         &  1235 \\
12  &  -                   &   -    &  -                   &   -   & Parking Lot 1   &  1233 \\
13  &  -                   &   -    &  -                   &   -   & Parking Lot 2   &  469  \\
14  &  -                   &   -    &  -                   &   -   & Tennis Court    &  428  \\
15  &  -                   &   -    &  -                   &   -   & Running Track   &  660  \\
\hline
    & Total                &  42776 & Total                & 7456   & Total           & 15029 \\
\hline
\end{tabular}
\end{table*}

\begin{table}
\caption{The distribution information of training, validation and test sets}
\label{tab: sample proportion}
    \centering
    \resizebox{0.47\textwidth}{11mm}{ 
    \begin{tabular}{c|c|c|c|c|c}
    \hline
        Setting & Dataset & Training & Validation & Test & Training\% \\ \hline
         & Pavia U & 180 & 90 & 42506 & 0.42\% \\ 
        20 pixels/class & Pavia C & 180 & 90 & 7186 & 2.41\% \\ 
         & Houston U & 300 & 150 & 14579 & 2.00\% \\ \hline
         & Pavia U & 270 & 90 & 42416 & 0.63\% \\ 
         30 pixels/class& Pavia C & 270 & 90 & 7096 & 3.62\% \\ 
         & Houston U & 450 & 150 & 14429 & 2.99\% \\ \hline
    \end{tabular}}
    \vspace{-0.2cm}
\end{table}

\subsection{Data Description}

To evaluate the effectiveness of the proposed NAS algorithm, we conduct comparison experiments on three representative HSI datasets, namely Pavia University, Pavia Center and Houston University. In turn, the false color composites and ground truth maps of these three HSIs are presented in Fig.\ref{fig:PU}-Fig.\ref{fig:HU}. The corresponding sample distribution information is listed in Table~\ref{tab: HU}. 

Pavia University and Pavia Center were captured by the ROSIS-3 sensor in 2001 during a flight campaign over Pavia, Nothern Italy. Due to low SNR, some frequency bands were removed. The remaining 103 channels are used for classification. These datasets have the same geometric resolution, that is  1.3 meters. Each dataset covers nine different land cover categories. Part of the categories are overlapped. Please find more details in Fig.\ref{fig:PU} and Fig.\ref{fig:PC}. Pavia University consists of 610 × 340 pixels and Pavia Centre covers 1096×715 pixels.

The Houston University was captured by the ITRES-CASI 1500 hyperspectral Imager over the University of Houston campus and the neighboring urban area. Compared with the aforementioned two datasets, Houston University has lower spatial resolution but much higher spectral resolution. Its spatial resolution is 2.5 $m$ and it contains 144 spectral bands, covering the wavelength range of 360–1050 µm. This dataset also covers a wider area and more abundant land cover objects. In specific, the Houston University dataset consists of 349 × 1905 pixels and includes 15 land-cover classes of interest.

\subsection{Experiment Design}

In order to validate the effectiveness of the proposed algorithm, we conduct experiments in two different settings. In setting one, 20 and 10 labeled pixels are randomly extracted from each class to build a training set and validation set. The rest is used as a test set. In setting two, the number of training samples of each class is increased to 30. Others keep the same with that in setting one. More details about the sample distribution are listed in Table~\ref{tab: sample proportion}. To ensure the fairness and stability of the comparison, we repeat each experiment five times and take the average values as the final results.

\subsection{Implementation Details}

Similar to 3D-ANAS~\cite{zhang20213d}, the proposed method also has two optimizing stages and one inference stage. In this section, we introduce the different settings in the aforementioned stages on three different datasets. For brevity, the settings that are consistent with the baseline 3D-ANAS would not be mentioned here.

\begin{figure*}[htbp]
	\begin{center}
		\includegraphics[width=0.78\linewidth]{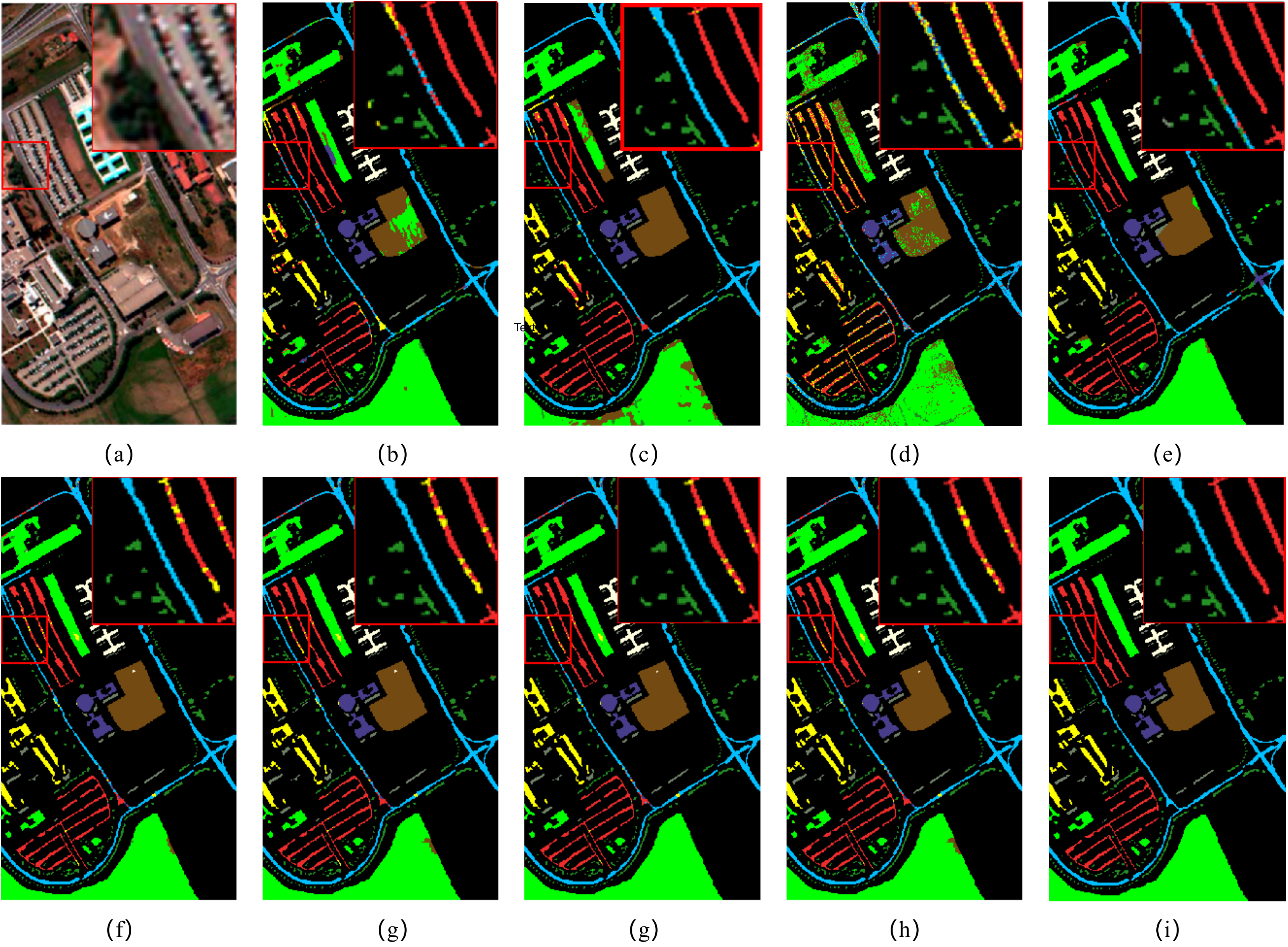}
	\end{center}
	\vspace{-0.4cm}
	\caption{ Comparison experimental results on Pavia University using 30 training samples in each class. (a) False color composite; (b) 3D-LWNet, OA=89.25\%; (c) SSRN, OA=91.95; (d) 1-D Auto-CNN, OA=81.75\%; (e) 3-D Auto-CNN, OA=93.36\%; (f) 3D-ANAS, OA=98.05\%; (g) Hy-NAS, OA=98.55\%; (h) HyT-NAS, OA=99.18\%; (i) HyT-NAS+OV, OA=99.52\%; (j) Ground truth map; }
	\label{fig:PU_result}
\end{figure*}

\newcolumntype{C}[1]{>{\centering\let\newline\\\arraybackslash\hspace{0pt}}m{#1}}
\begin{table}[ht]
\setlength{\arraycolsep}{-0.1pt} 
\setlength{\abovecaptionskip}{0.cm}
\setlength{\belowcaptionskip}{-0.cm}
\renewcommand{\arraystretch}{1.3}
\caption{Comparison Experimental Results on Pavia University Using 20 Training Samples Each Class}
\label{tab: PaviaU_20}
\centering
\begin{threeparttable}
\resizebox{3.5in}{!}{
\begin{tabular}{C{0.77cm}|C{0.7cm}C{0.65cm}C{0.65cm}C{0.65cm}C{0.65cm}|C{0.65cm}C{0.68cm}C{0.65cm}C{0.8cm}C{0.8cm}}
\hline
Models & 3D-LWNet & SSRN & 1-D Auto-CNN & 3-D Auto-CNN & 3D-ANAS\dag  & Hy-NAS & HyT-NAS     & HyT-NAS +OV\\
\hline
    
1      &  82.43  & 99.54 &  69.69         &  88.24             &    92.72          &   97.77           &   \textbf{98.88}      &98.79\\  
2      &  84.76  & 99.31  &  76.37         &  90.72             &    96.18          &   98.88  &   97.86                        &\textbf{99.16}\\ 
3      &  76.88  & 94.36 &  73.43         &  92.11             &    97.32          &   98.07           &   \textbf{98.44}      &98.12\\ 
4      &  91.45 & 94.95   &  90.2          &  81.27             &    95.49          &   95.81           &    97.45              &\textbf{98.06}\\ 
5      &  96.23  & 99.25  &  96.54         &  93.12             &    \textbf{100}   &   99.92           & 99.31                 &99.77\\ 
6      &  92.50  & 71.76  &  75.48         &  98.47             &    96.89          &   94.48           &   \textbf{99.56}      &98.92\\ 
7      &  93.56 & 73.64  &  88.83         &  96.14             &    97.19          &   99.92           &  \textbf{100}         &\textbf{100}\\ 
8      &  96.01  & 86.27  &  77.59         &  \textbf{96.84}    &    93.64          &   92.03           &    95.19              &96.44\\ 
9      &  89.16 & 98.72   &  96.66         &  79.62             &    \textbf{100}   &   \textbf{100}    &  \textbf{100}         &\textbf{100}\\ 
\hline
OA     &  87.10  & 91.99  &  77.65         &  91.16      &    95.74    &97.43    &98.03   &\textbf{98.77}\\ 
AA     &  89.22  & 90.87   &  82.75         &  90.72      &    96.60    &97.43    &98.41  &\textbf{98.81}\\ 
K      &  83.35  & 89.60   &  71.51         &  88.51      &    94.37    &96.59    &97.39  & \textbf{98.37}\\ 
\hline
\end{tabular}
}
\end{threeparttable}
\end{table}

\begin{table}[!t]
\setlength{\abovecaptionskip}{0.cm}
\setlength{\belowcaptionskip}{-0.cm}
\renewcommand{\arraystretch}{1.3}
\caption{Comparison Experimental Results on Pavia University Using 30 Training Samples Each Class}
\label{tab: PaviaU_30}
\centering
\begin{threeparttable}
\resizebox{3.5in}{!}{
\begin{tabular}{C{0.77cm}|C{0.7cm}C{0.65cm}C{0.65cm}C{0.65cm}C{0.65cm}|C{0.65cm}C{0.68cm}C{0.65cm}C{0.8cm}C{0.8cm}}
\hline
Models & 3D-LWNet & SSRN & 1-D Auto-CNN & 3-D Auto-CNN & 3D-ANAS\dag  & Hy-NAS & HyT-NAS &HyT-NAS +OV\\
1 &  82.32   & 99.90   &  76.43         & 89.70       & 95.24          &   94.43           &   \textbf{99.07}  &99.03\\  
2 &  88.84  & 99.53    &  81.9          & 97.92       & 98.16          &   99.74           &   99.26           &\textbf{99.80}\\ 
3 &  83.74  & 88.26    &  74.04         & 92.31       & 97.44          &   99.32           &   99.17           &\textbf{99.81}\\ 
4 &  94.11  & 93.16     &  93.38         & 71.22       & 98.52          &   \textbf{98.52}  &   97.45           &97.82\\ 
5 &  96.90  & 100    &  96.13         & 95.12       & 99.77          &   \textbf{100}    &   \textbf{100}    & \textbf{100}\\ 
6 &  92.10  & 66.97    &  81.81         & 96.96       & 99.46          &   \textbf{100}    &   99.74           & \textbf{100}\\ 
7 &  95.46  & 99.69   &  88.23         & 95.99       & 99.92          &   \textbf{100}    &   \textbf{100}    &99.92\\ 
8 &  93.19  & 89.77    &  74.05         & 94.98       & 98.85          &   96.19           &   98.90           &\textbf{99.18}\\ 
9 &  92.43  & 99.24    & 95.65          & 80.64       & 99.89          &   \textbf{100}    &   \textbf{100}    & \textbf{100}\\ 
\hline
OA   &  89.25 & 91.95  &  81.75         & 93.36       & 98.05      & 98.55    &99.18      &\textbf{99.52}      \\ 
AA   &  91.01  & 92.95 &  84.62         & 90.54       & 98.58      & 98.69    &99.29      &\textbf{99.51}      \\ 
K    &  86.05  & 89.59 &  76.55         & 91.50       & 97.42      & 98.08    &98.92      &\textbf{99.37}      \\ 
\hline
\end{tabular}
}

\end{threeparttable}
\end{table}

\begin{figure*}[ht]
	\begin{center}
		\includegraphics[width=0.85\linewidth]{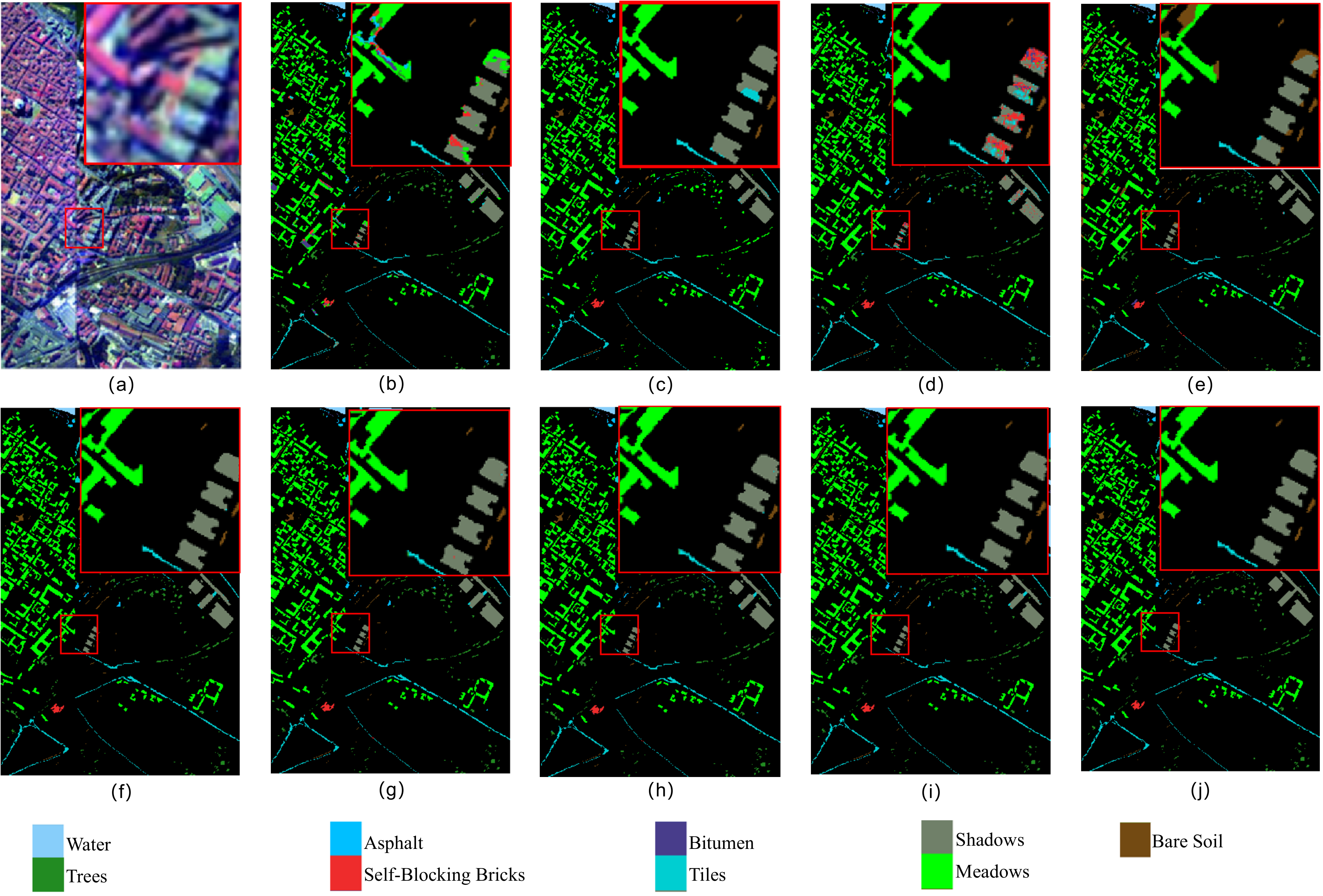}
	\end{center}
	\caption{Comparison experimental results on Pavia Centre using 30 training samples each class. (a) False color composite; (b) 3D-LWNet, OA=94.22\%; (c)SSRN, OA=98.72; (d)1-D Auto-CNN, OA=96.79\%; (e) 3-D Auto-CNN, OA=96.99\%; (f) 3D-ANAS, OA=99.24\%; (g) Hy-NAS, OA=99.33\%; (h) HyT-NAS, OA=99.44\%; (i) HyT-NAS+OV, OA=99.49\%; (j) Ground truth map; }
	\label{fig:PC_result}
\end{figure*}

\vspace{5 pt}
\noindent\textbf{Searching:} For three different datasets, we construct three different super nets, which share the same outline structure. Specifically, in the outer structure, each super net consists of four layers of super cells and each layer is made up with two different super cells, space dominated super cell and spectrum dominated super cell. In the inner structure, each cell has a sequence of three nodes. The entire searching process is carried out on a NVIDIA V100 card with 32G memory. 
For Pavia University and Pavia Centre, we crop the patches with spatial resolution of 24×24 as searching samples, and the batch size is set to 6. On Houston University, the crop size of patches is set to 14x14, and the batch size is 5. On all three datasets, the Adam optimizer with both learning rate and weight attenuation of 0.001 is used to optimize the architecture parameters ($\alpha$, $\beta$ and $\omega$). The standard SGD optimizer is applied to update the super net parameters (learnable kernels in candidate operations), where momentum and weight decay are set to 0.9 and 0.0003, respectively. The learning rate decays from 0.025 to 0.001 according to the cosine annealing strategy.  For Pavia University and Pavia Centre, the first 15 epochs are the warm-up stage, in which we only optimize super net parameters. Because Houston University is more challenging, we set 30 epochs for warming up. After the warming-up stage, we alternately update the architecture parameters and super network parameters in each iteration.

\begin{table}
\setlength{\abovecaptionskip}{0.cm}
\setlength{\belowcaptionskip}{-0.cm}
\renewcommand{\arraystretch}{1.3}
\caption{Comparison Experimental Results on Pavia Centre Using 20 Training Samples Each Class}
\label{tab: Pavia_20}
\centering
\begin{threeparttable}
\resizebox{3.5in}{!}{
\begin{tabular}{C{0.77cm}|C{0.7cm}C{0.65cm}C{0.65cm}C{0.65cm}C{0.65cm}|C{0.68cm}C{0.65cm}C{0.8cm}C{0.7cm}c}
\hline
Models & 3D-LWNet & SSRN & 1-D Auto-CNN & 3-D Auto-CNN & 3D-ANAS\dag & Hy-NAS & HyT-NAS     &HyT-NAS +OV\\
1    &  99.61  & \textbf{100}  &   99.81       &   99.56        &  99.61      &  99.62    &  99.71   &99.69\\  
2    &  91.85  & 98.68  &   80.9        &   87.79        &  93.16      &   95.18   &   \textbf{96.61}  &96.43\\ 
3    &  86.77  & 87.39  &   89.04       &   82.98        &  \textbf{96.67}      &   93.43   &   89.77  &89.28\\ 
4    &  92.85  & 87.79  &   73.53       &   98.85        &  99.81      &   96.80   &   \textbf{100}    &\textbf{100}\\ 
5    &  95.53  & 99.75  &   90.05       &   95.83        &  98.14      &   99.80   &   99.99           &\textbf{100}\\ 
6    &   80.97 & 88.58  &   95.53       &   93.22        &  98.70      &   \textbf{99.23}   &   97.06  &97.97\\ 
7	 &   85.66 & 99.28  &   84.23       &   94.94        &  94.72      &   95.48   &   98.85           &\textbf{99.21}\\ 
8    &   85.06 & \textbf{100}   &   98.17       &   95.12        &  99.89      &   99.77   &   99.95           &99.96\\ 
9    &   91.34 & 97.79 &   98.89       &   85.29        &  99.82      &   \textbf{100}     & 99.75    &99.89\\ 
\hline
OA   &   92.42  & 98.51 &  96.18        &   96.25        & 98.95       &   99.05   &   99.23  &\textbf{99.28}\\ 
AA   &   89.96  & 95.47 &  90.02        &   92.62        & 97.84       &   97.70   &   97.98  &\textbf{98.05}\\ 
K    &   89.41  & 97.89 &  94.6         &   94.71        & 98.51       &   98.65   &   98.81  &\textbf{98.98}\\ 
\hline
\end{tabular}
}

\end{threeparttable}
\end{table}

\noindent\textbf{Grafted network optimization:} We crop patches with spatial resolution 32×32 to train the final grafted network. Random cropping, flipping, and rotation are introduced as data enhancement strategies. Batch sizes on Pavia University and Pavia Centre are set to 12. Batch size on Houston University is set to 16. At this stage, we use the SGD optimizer. The initial learning rate is set to 0.1, decayed according to the poly learning rate policy with power of 0.9 ($lr=init\_lr\cdot({1-\frac{iter}{max\_iter}}^{power})$). The performance of the network is validated every 100 iterations.

\begin{table}
\setlength{\abovecaptionskip}{0.cm}
\setlength{\belowcaptionskip}{-0.cm}
\renewcommand{\arraystretch}{1.3}
\caption{Comparison Experimental Results on Pavia Centre Using 30 Training Samples Each Class}
\label{tab: Pavia_30}
\centering
\begin{threeparttable}
\resizebox{3.5in}{!}{
\begin{tabular}{C{0.77cm}|C{0.7cm}C{0.65cm}C{0.65cm}C{0.65cm}C{0.65cm}|C{0.65cm}C{0.68cm}C{0.65cm}C{0.8cm}C{0.8cm}}
\hline
Models & 3D-LWNet & SSRN & 1-D Auto-CNN & 3-D Auto-CNN & 3D-ANAS\dag & Hy-NAS & HyT-NAS     &HyT-NAS +OV\\
1   &   99.52   & 100  &     99.76       &   99.78     &   \textbf{100.0}        &\textbf{100.0}     &99.99              &\textbf{100.0}\\  
2   &   93.92   & 99.34  &     87.9        &   92.48     &   95.83                 &96.75     &96.40                       &\textbf{96.98}\\ 
3   &   89.26   & 86.05  &     91.16       &   85.81     &   93.34                 &95.28              &\textbf{96.43}     &96.20\\ 
4   &   91.1    & 79.70  &     79.72       &   97.32     &   97.77                 &99.85              &99.96              &\textbf{100.0}\\ 
5   &   96.09   & 99.93   &     92.1        &   97.02     &   98.36                 &96.82              &99.95              &\textbf{99.97}\\ 
6   &   90.73   & 93.62  &     96.47       &   95.91     &   99.57                 &98.97              &99.69              &\textbf{99.90}\\ 
7	&   93.24   & 99.51   &     85.43       &   95.29     &   97.97                 &97.92              &\textbf{98.65}     &98.58\\ 
8   &   87.32   & \textbf{99.99}   &     97.88       &   95.13     &   99.40   &  99.69     &99.27              &99.31\\ 
9   &   93.7    & 99.23   &     98.58       &   91.84     &   \textbf{100.0}        &99.89              &99.79              &\textbf{100.0}\\ 
\hline
OA  &    94.22  & 98.72    &    96.79       &   96.99     &   99.24  &   99.33   &   99.44      &\textbf{99.49}\\ 
AA  &    92.76  & 95.26  &    92.11       &   94.51     &   98.03  &   98.34   &   98.90      &\textbf{98.99}\\ 
K   &    91.92  & 98.18  &    95.47       &    95.76    &   98.92  &   99.05   &   99.20      &\textbf{99.28}\\ 
\hline
\end{tabular}
}
\end{threeparttable}
\end{table}

\noindent\textbf{Inference:} For the grafted framework based on hybrid CNN and transformer, we introduced an overlap inference strategy (OV) to further improve the performance. Specifically, we use a sliding window to crop small blocks (the stride is half of the window size), and input the cropped blocks into the compact network. The average result of the overlapping area is considered as the final prediction result. As the number of tokens in our transformer module is fixed, the multi-scale verification method (MS) is not adopted here. While, using OV strategy alone already achieves promising performance. The structure we designed requires the input sequence to be a fixed length. Therefore, the image blocks should be on the same scale during training and verification. Relaxing this restriction is considered as one of our future work.

\begin{table}[!t]
\setlength{\abovecaptionskip}{0.cm}
\setlength{\belowcaptionskip}{-0.cm}
\renewcommand{\arraystretch}{1.3}
\caption{Comparison Experimental Results on Houston University Using 20 Training Samples Each Class}
\label{tab: Houston_20}
\centering
\begin{threeparttable}
\resizebox{3.5in}{!}{
\begin{tabular}{C{0.77cm}|C{0.7cm}C{0.65cm}C{0.65cm}C{0.65cm}C{0.68cm}|C{0.65cm}C{0.65cm}C{0.65cm}}
\hline
Models & 3D-LWNet & SSRN & 1-D Auto-CNN & 3-D Auto-CNN & 3D-ANAS\dag & Hy-NAS & HyT-NAS     &HyT-NAS +OV\\
1   &   79.16     & 69.69 &   69.98             &   84.01               &   77.56           &   83.62       &   \textbf{87.37}      &86.00\\  
2   &   71.61     & 96.81 &   60.19             &   85.65               &   82.27           &   79.66       &   88.30              &\textbf{88.48}\\ 
3   &   94.4      & 97.81 &   77.30             &   93.86               &   86.96           &   82.01       &   \textbf{87.67}      &86.96\\ 
4   &   74.25     & 85.79 &   49.02             &   66.77               &   \textbf{82.78}  &   70.92       &   71.76               &72.41\\ 
5   &   90.37     & 95.82 &   83.09             &   93.83               &   91.17           &   93.15       &   \textbf{97.00}      &96.70\\ 
6   &   84.92     & 81.71 &   50.46             &   80.43               &   96.27           &   94.58       &   98.95               &\textbf{98.98}\\ 
7   &   \textbf{84.73} & 64.64   &   30.93     &   72.21               &   72.29           &   80.45       &   83.39               &\textbf{84.09}\\ 
8   &   52.01     & 96.53  &  50.53             &   70.96               &   60.79           &  78.83        &    80.15              &\textbf{80.56}\\ 
9   &   70.47     & 76.11 &   46.58             &   \textbf{71.18}      &   69.72           &   68.74       &   70.38               &\textbf{72.67}\\ 
10  &   92.15     & 83.35 &   69.73             &   96.43               &   86.55           &   92.23       &   97.14               &\textbf{99.00}\\ 
11  &   89.39     &  85.10 &   50.06            &   \textbf{93.73}      &   91.95           &   86.14       &   92.72               &91.70\\ 
12  &   60.04     &  74.15 &  63.44            &   87.95               &   84.37           &   85.12       &   \textbf{88.60}      &86.95\\ 
13  &   86.99     &  84.25 &   53.35            &   84.90               &   \textbf{97.27}  &   91.57       &   95.57               &95.44\\ 
14  &   92.99     & 93.14  &  76.73            &   92.99               &   99.25           &   98.24       &   \textbf{100}        &\textbf{100}\\ 
15  &   92.83     & 96.01  & 54.00             &   88.30               &   91.43           &   94.13       &   92.58      &\textbf{92.70}\\ 
\hline
OA  &    78.95    & 82.55 & 58.35        &   83.37     &  82.10     &   83.39     &   86.97      & \textbf{87.11}\\ 
AA  &    81.09    & 85.39 &  59.03        &   84.21     &  84.71     &   85.29     &   88.77      & \textbf{88.84}\\ 
K   &    77.32    & 81.13 &  55.18        &   82.07     &  80.67     &   82.06     &   85.92      & \textbf{86.07}\\
\hline
\end{tabular}
}
\end{threeparttable}
\end{table}

\subsection{Comparison with state-of-the-art methods}

\begin{figure*}[htbp]
	\begin{center}
		\includegraphics[width=0.85\linewidth]{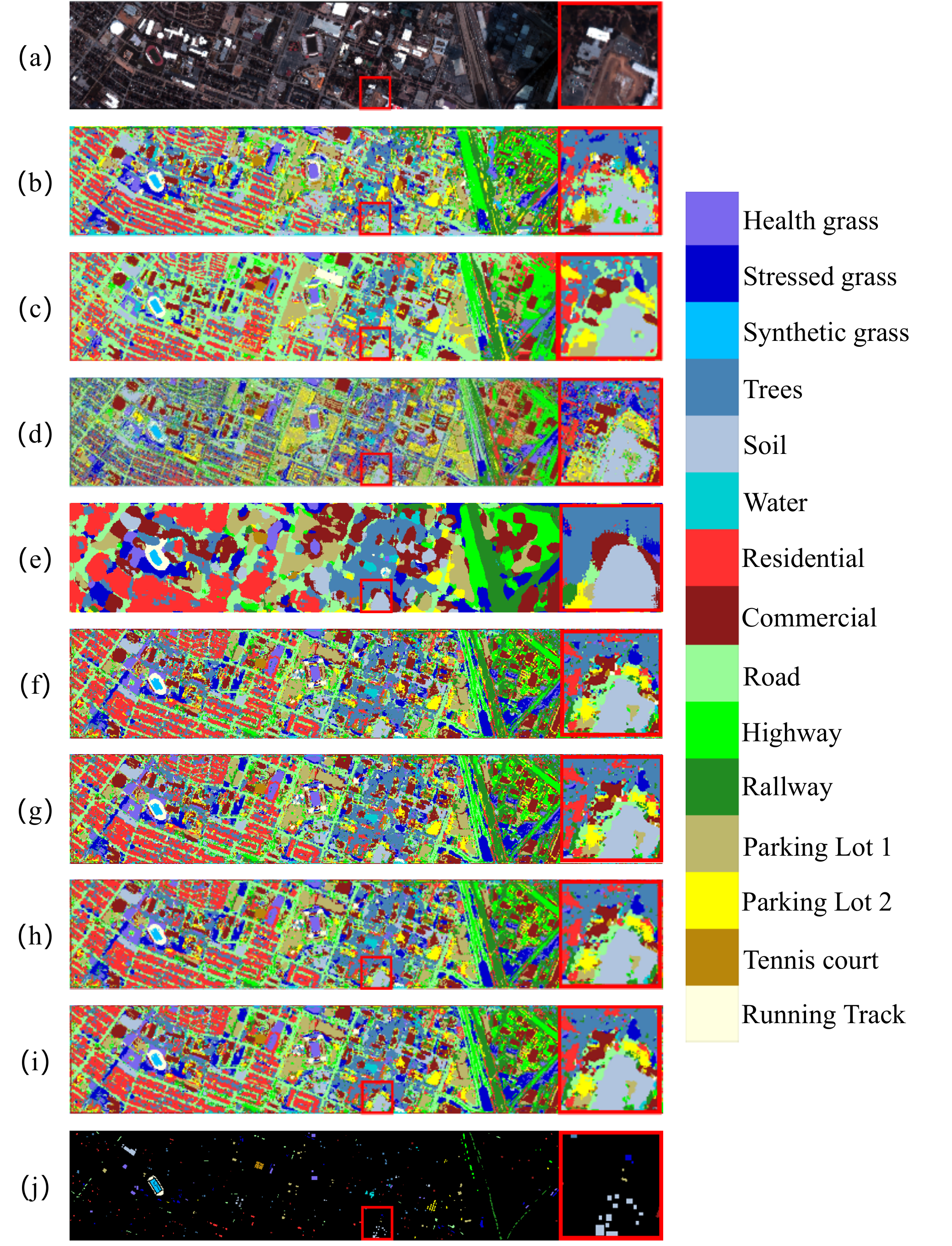}
	\end{center}
	\caption{ Comparison experimental results on Houston University using 30 training samples in each class. (a) False color composite; (b) 3D-LWNet,
OA=84.17\%; (c) SSRN, OA=86.60; (d) 1-D Auto-CNN, OA=60.36\%; (e) 3-D Auto-CNN, OA=84.45\%; (f) 3D-ANAS, OA=85.81\%; (g) Hy-NAS+MS, OA=86.22\%; (h) HyT-NAS, OA=90.04\%; (i)  HyT-NAS+OV, OA=91.14\%; (g) Ground truth map;}
\label{fig:HU_result}
\end{figure*}

In this section, we compare the proposed Hy-NAS and HyT-NAS with other four recent CNN-based HSI classification methods. The codes for all comparison methods are derived from the official codes: 3D\-LWNet\footnote{\url{https://github.com/hkzhang91/LWNet}}, 1-D Auto-CNN and 3-D Auto-CNN\footnote{\url{https://github.com/YushiChen/Auto-CNN-HSI-Classification}} and 3D-ANAS\footnote{\url{https://github.com/hkzhang91/3D-ANAS}}. TABLE~\ref{tab: PaviaU_20}-\ref{tab: Houston_30} list the results of the comparative experiment, and Figs.\ref{fig:PU_result}-\ref{fig:HU_result} show the corresponding visual results. Here, we do not compare the inference speeds, as these methods are implemented with different frameworks, which may introduce biases. The inference of pixel-to-pixel framework has higher efficiency than that of patch-to-pixel framework, because the former framework does not have repeat operations as explained in~\cite{zhang20213d}. The method proposed in this paper adopts a pixel-to-pixel framework and inherits the advantage in inference efficiency. 

The performance on the Pavia University is listed in TABLE~\ref{tab: PaviaU_20} and TABLE \ref{tab: PaviaU_30}. The corresponding visual comparison results are shown in Fig.\ref{fig:PU_result}. From the comparison results, we can draw the following conclusions: 1) Compared with the method based on 1D CNN, the method based on 3D CNN usually gains better performance. Because jointly using the spectral and spatial information is beneficial to improve the classification accuracy. 
 
Compared to 3D-ANAS~\footnote{For fairness, we rerun the code of 3D-ANAS on the same device (with nvidia V100 GPUs) with the proposed methods and reported the results as 3D-ANAS\dag. For 3D-ANAS\dag, the performance is a little bit different from that in the orignial paper, we conjecture this is caused by the re-implemention on different devices.}, the proposed method adopts hybrid search space, which improves the flexibility in processing spectral and spatial with different operations, achieving higher classification accuracy. 2) After grafting the transformer structure, the proposed HyT-NAS achieves better performance than other auto-designed methods. For example, HyT-NAS achieves 98.03\% OA, 98.41\% AA, and 97.39\% K when 20 training samples are extracted from each class, which are 2.29\%, 1.81\%, and 3.02\% higher than 3D-ANAS, respectively. 3) The overlap inference enhancement strategy can further improve the performance. As shown in Table \ref{tab: PaviaU_20}, using OV increases OA, AA, and K by 0.74\%, 0.40\%, and 0.98\%, respectively.

\begin{table}[!t]
\setlength{\abovecaptionskip}{0.cm}
\setlength{\belowcaptionskip}{-0.cm}
\renewcommand{\arraystretch}{1.3}
\caption{Comparison Experimental Results on Houston University Using 30 Training Samples Each Class}
\label{tab: Houston_30}
\centering
\begin{threeparttable}
\resizebox{3.5in}{!}{
\begin{tabular}{C{0.77cm}|C{0.7cm}C{0.65cm}C{0.65cm}C{0.65cm}C{0.68cm}|C{0.65cm}C{0.65cm}C{0.65cm}}
\hline
Models & 3D-LWNet & SSRN & 1-D Auto-CNN & 3-D Auto-CNN & 3D-ANAS\dag & Hy-NAS & HyT-NAS &HyT-NAS +OV\\
1   &   84.81  &  90.91 & 72.25           &   87.50                &   \textbf{90.01}  &    \textbf{90.01}       &   78.94     &80.59\\  
2   &   80.22  &  98.61 &   63.96            &   77.91               &  79.49            &   85.75       &    92.42              &\textbf{92.75}\\ 
3   &   93.45  &  97.75 & 76.61            &   92.74               &   98.78           &   85.08       &   98.17               &\textbf{99.85}\\ 
4   &   79.74  &  82.11 & 51.29            &   72.65               &  85.22            &   86.63       &    91.53              &\textbf{91.94}\\ 
5   &   90.90   & 91.51 &   82.25           &   96.14               &  99.00            &    \textbf{99.50}       &   96.84     &98.34\\ 
6   &   81.44  &  65.83 &   55.32           &   84.86               &  91.93            &   97.89       &    \textbf{99.30}     &\textbf{99.30}\\ 
7   &   87.83  &  81.52 &    34.20            &   73.03               &  66.53            &   76.30       &   83.06               &\textbf{91.43}\\ 
8   &   63.69  &  98.15 &    62.11           &   76.64               &  75.50            &   72.43       &    77.24              &\textbf{77.99}\\ 
9   &   77.50   & 67.71 &   49.84           &   71.10               &  81.93            &   74.34       &    86.72             &\textbf{88.45}\\ 
10  &    93.26 &  91.49 &    64.48           &   96.67               &  90.14            &   89.81       &    \textbf{97.72}     &96.88\\ 
11  &    91.04 &  93.38 &  50.23           &   92.31               &  87.36            &   87.03       &    94.73              &\textbf{96.23}\\ 
12  &    79.40  & 90.44 &   62.47           &   91.48               &  89.52            &   87.18       &   92.20              &\textbf{92.29}\\ 
13  &    90.41 &  69.84 &     50.92           &   85.8                &  88.81            &   94.64       &    98.60              &\textbf{99.07}\\ 
14  &    90.65 &  84.09 &    76.07           &   90.65               &   \textbf{99.23}  &   96.13       &   96.91               &96.39\\ 
15  &    92.47 &  97.27 &  62.94            &    88.85              &  85.97            &   96.29       &   94.03               &\textbf{98.39}\\
\hline
OA  &    84.17  & 86.60 &  60.36        &   84.45     &    85.81     &   86.22     &    90.04     &\textbf{91.14}\\ 
AA  &    85.12  & 86.71 &   61.00       &   85.22     &    87.29     &   87.93     &    91.89     &\textbf{92.66}\\ 
K   &    82.96  & 85.53 &   57.35       &   83.25     &    84.66     &   85.10     &    89.64     &\textbf{90.42}\\ 
\hline
\end{tabular}
}
\end{threeparttable}
\end{table}

\begin{figure*}[htbp]
	\begin{center}
		\includegraphics[width=0.9\linewidth]{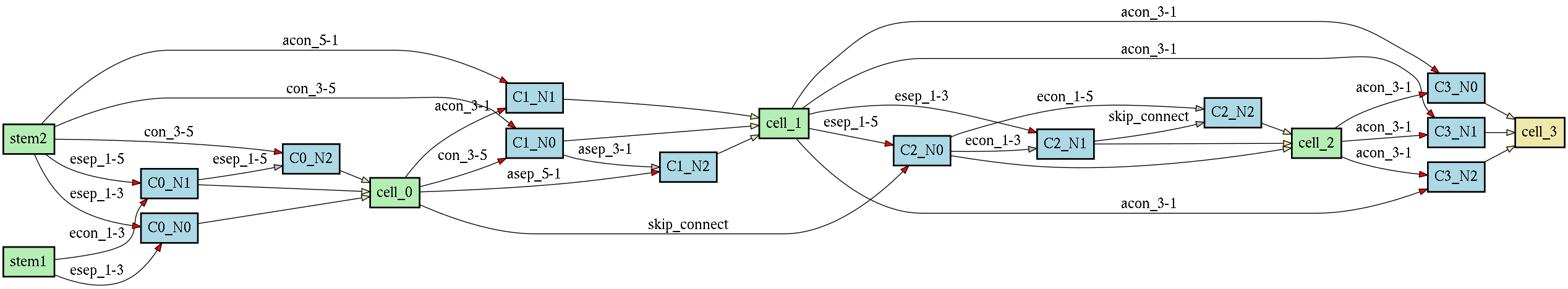}
		\includegraphics[width=0.9\linewidth]{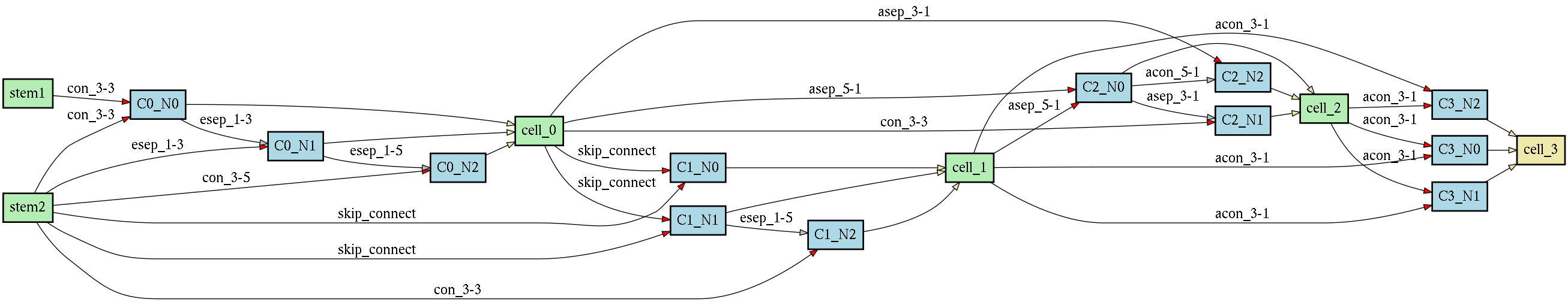}
		\includegraphics[width=0.9\linewidth]{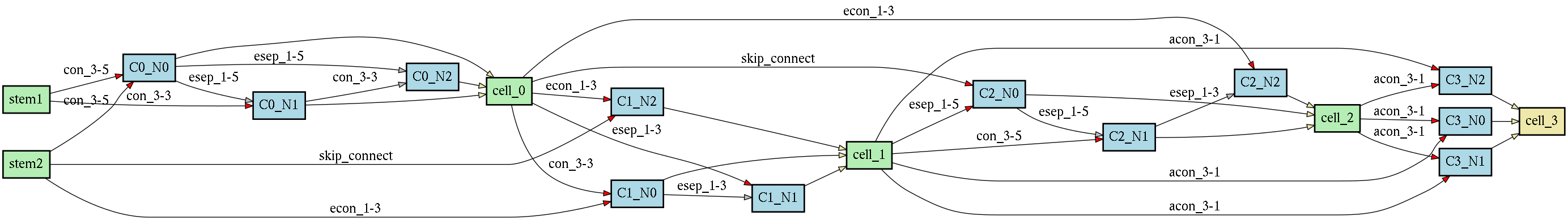}
	\end{center}
	\vspace{-0.3cm}
	\caption{The final architectures found for three datasets. From up to down are: Pavia University; Pavia Centre and Houston University.}
	\label{fig:fig10}
\end{figure*}

To save space, we only present the visual results using 30 training samples per class in Fig.\ref{fig:PU_result}. In order to clearly illustrate the difference, we placed a partially enlarged patch in the upper right corner of each result map. It can be easily found from the partially enlarged patch that there are fewer misclassified pixels in the results of a series of HyT-NAS. Some Asphalt pixels (class 1, cyan) are incorrectly classified as Self-Blocking Bricks (class 8, red) by 3D-LWNet and 3-D Auto-CNN. A lot of pixels belonging to Self-Blocking Bricks are incorrectly classified as Meadows (Class 2, green) by 3D-ANAS. But in the results of a series of HyT-NAS, all pixels belonging to Asphalt and Self-Blocking Bricks are correctly classified.

TABLE \ref{tab: Pavia_20} and \ref{tab: Pavia_30} collects the comparison results on Pavia centre, and Fig.\ref{fig:PC_result} shows the visual results of qualitative analysis. Compared with the results on Pavia University, the accuracy of these seven methods all improved to certain extents and the proposed HyT-NAS still attains the best performance. Observing from Fig.\ref{fig:PC_result}, the number of bitumen pixels that a series of HyT-NAS approaches incorrectly classified into self-blocking Bricks is significantly less than that of other methods. Although Hy-NAS with only improved hybrid spatial-spectrum search space still makes some false predictions on the bitumen class, the introduction of the transformer finally handles the problems very well.

\begin{table}[!t]
\setlength{\abovecaptionskip}{0.cm}
\setlength{\belowcaptionskip}{-0.cm}
\renewcommand{\arraystretch}{1.2}
\setlength{\tabcolsep 4pt}
\caption{Comparisons between different search spaces on Houston University}
\label{tab: ablation}
\centering
\begin{tabular}{ccc|ccc}
\hline
Search Space   & Model Size    & Transformer   & OA    & AA    & K \\
\hline
Spectral  &   1.41 MB           &  \ding{55}               & 83.04            &   85.25       & 81.68\\ 
Spatial &    1.48 MB           &  \ding{55}               &  84.85            &   86.90       & 83.63\\ 
Spectral + Spatial  &   1.44 MB           &  \ding{55}               & 86.22            &   87.93       & 85.10\\ 
\hline
Spectral  &   9.98 MB          &  \ding{51}               &       88.18     &  89.75      & 87.22  \\
Spatial   &    10.04 MB          &  \ding{51}               &      89.53        &   90.96       &  88.67  \\ 
Spectral + Spatial   &   10.00 MB           &  \ding{51}               & 90.04            &   91.89       & 89.64\\ 
\hline
\end{tabular}
\vspace{-0.3cm}
\end{table}

The comparison results on Houston University are shown in Table \ref{tab: Houston_20} and \ref{tab: Houston_30} and Fig.\ref{fig:HU_result}. Compared with the first two datasets, the Houston University contains more spectral bands and more object categories. Therefore, the classification accuracy of all methods on this dataset is relatively low. The classification performance of different methods is quite different. As shown in Fig.\ref{fig:HU_result}, the result map of 1D Auto-CNN clearly shows the structural outlines of different buildings. For example, the dark red part of the partially enlarged area (commercial, level 8). But many misclassified pixels are distributed throughout the result image and look like salt and pepper noise, resulting in relatively poor visual effects. On the contrary, 3D Auto-CNN showed very smooth results, in which the outline of the structure was almost lost. 3D-ANAS and HyT-NAS have kept a relatively good balance between displaying good visual effects and maintaining the contour structure, and gained better performance than other algorithms. As shown in the enlarged image in Fig.\ref{fig:HU_result}, 3D-ANAS misclassifies some pixels classified as land into stressed grass and highway, while HyT-NAS has very few misclassified pixels. From the TABLE \ref{tab: Houston_20} and \ref{tab: Houston_30}, it is obvious that the results of HyT-NAS are better than those of 3D-ANAS regardless of whether the training samples are 20 or 30. Specifically, when there are 20 training samples for each class, the OA, AA, and K of HyT-NAS are 86.97\%, 88.77\%, and 85.92\%, respectively, which are significantly higher than that of 3D-ANAS. When the training samples of each class increase to 30, the advantages of HyT-NAS and 3D-ANAS are more obvious. increasing by 4.23\%, 4.60\%, and 4.98\% on OA, AA, and K respectively.

\subsection{Ablation study}

HSI has different spatial and spectral resolutions. During the searching stage, different layers tend to select different types of cells. We speculate that merely maintaining a space dominated cell or a spectrum dominated cell would affect the performance of the algorithm, although both kinds of cells contain the 3D convolution. Here, ablative experiments are conducted to verify the effectiveness of hybrid search space. Besides, we also compared the classification accuracy of the model with and without the transformer unit. The experiments are carried out on the most challenging dataset Houston University with 30 training samples per class. 

As shown in TABLE ~\ref{tab: ablation}, the proposed method with hybrid search space achieves the highest accuracy. When only the spectral search space is retained in the model, the classification accuracy is the lowest. Importing different types of cells can dig out the spatial and spectral information jointly and freely in HSI classification tasks. In addition, the introduction of the transformer has brought about a 5\% improvement in accuracy in all different search space settings. This illustrates the importance of fully mining the associated information between pixels in the classification of HSIs.

\subsection{Architecture analyze}
In this section, the architectures searched by HyT-NAS are shown in Fig.\ref{fig:fig10} and analyzed. Since the three datasets have different spectral and spatial resolutions, and land covers, we searched for the architecture on each dataset separately. Although these three architectures are different in topology and operations, they also have some common characteristics:
 
1) 2D spatial convolution and 2D spectral convolution play important roles in the final selected operations. As introduced in Section III-B, the search space we construct for searching the internal topology includes not only 2D spatial convolution and 2D spectral convolution, but also 3D convolution. Even so, HyT-NAS tends to build a network with both 2D convolution operations and 3D convolution operations. In most cases, 2D convolution operations are the main operation and 3D convolution operations play the part of the complementary operation. The proportion of 2D convolution operations in the final network designed on Pavia centre and Pavia University are 41.67\% and 52.78\%, respectively. Under the architecture searched for Houston University, 2D convolution operations occupied 44.44\% of all operations. The proportions of 3D convolution operations are 13.89\%, 8.33\%, and 16.67\% on Pavia centre, Pavia University and Houston University.  This shows that although 3D convolution fits the data characteristics of HSIs, widely utilizing it as in traditional algorithms is not necessary. The 2D-3D mixed network architecture we searched has fewer parameters and higher parameter utilization compared with models under the same scale.

2) In the final network, 3D convolution operations are distributed from the beginning to the end. However, 2D spectral convolutions dominate in the shallow layer, while the number of 2D spatial convolutions is relatively large in the deep layer, as shown in Fig.\ref{fig:fig10}. In the architectures for the Pavia Centre and Pavia University, the spectral convolutions account for the majority in the first two layers, but the spatial convolutions account for the highest proportion in the last two layers. In the final architecture for Houston University, the first three layers are almost all spectral convolutions, and only the last layer of the network is mainly spatial convolution. In classic HSI classification networks such as SSRN~\cite{zhong2018spectral}, spectral convolution is always performed first, followed by the spatial convolution. Our experimental results are consistent with the manual design experience.

3) With the enrichment of spectral information, the proportion of spectral convolution in the final network gradually increases. In different HSIs, the richness of spectral information and spatial information are quite different.  For example, both Pavia University and Pavia Centre have only 102 bands, while Houston University has 144 bands. Traditional 3D convolution pays the same attention to spatial information and spectral information. The hybrid spatial-spectral search space we proposed can flexibly adjust the ratio of spatial and spectral convolutions freely according to the proportion of the spatial-spectral information of the data itself. As shown in the Fig.\ref{fig:fig10}, although the search space is exactly the same, 2D-spectral convolution and 3D convolution account for 33.33\% and 25.00\% of operations in Pavia University and Pavia centre, respectively. In the architecture for Houston University, this proportion has risen to 44.44\% . This may be because Houston University has a larger number of spectra bands and requires more spectral convolutions to extract rich spectral information. The experimental results prove that the hybrid search space can adapt well to the characteristics of different data.

\begin{figure}
	\begin{center}
		\includegraphics[width=1.0 \linewidth]{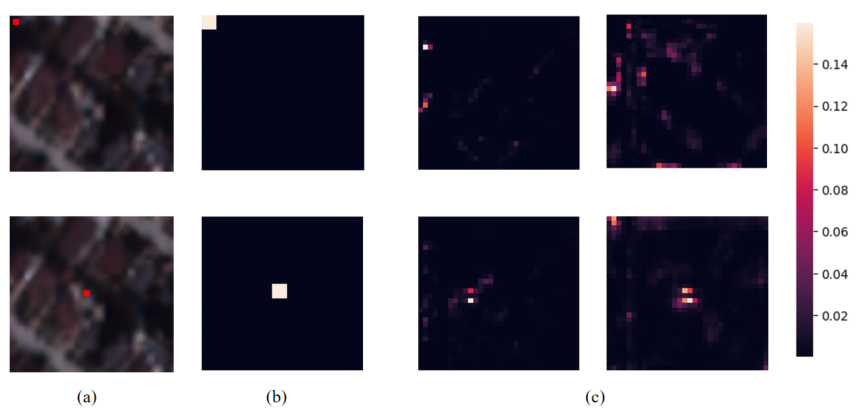}
	\end{center}
	\vspace{-0.3cm}
	\caption{Attention map analyse. (a) Input patch. Two pixels marked in red color are pixels to be classified ; (b) Region of interesting of CNN; (c) attention maps of different heads of transformer unit.}
	\label{fig:attention}
\end{figure}

\subsection{Attention map analyze}

In this section, we analyze the difference between using and not using the transformer unit on Houston University. A visualized result is presented in Fig.\ref{fig:attention}. After introducing the transformer unit, the receptive field of the feature map has expanded from a small range to a wider global one. In addition, the CNN structure pays the same level of attention to each pixel in a local receptive field, while the transformer unit can capture the relationship between global pixels in HSI adaptively. 

Fig.~\ref{fig:attention}(c) shows that the attention maps produced by different heads inside the transformer are also different. This observation indicates the effectiveness of multi-attention head mechanism in HSI classification tasks. Various heads inside the transformer focus on different types of information, according to the spatial position and neighborhood of the pixel itself. The multi-attention mechanism fuses the information of different heads, resulting in a more comprehensive and robust feature map.

\section{Conclusion}

In this paper, we have proposed an auto-designed HSI classification method based on the hybrid CNN-Transformer framework. The proposed HyT-NAS has been compared with other manual designed CNN based HSI classification methods (3D-LWNet), automatic design CNN based methods (1-D Auto-CNN, 3-D Auto-CNN and 3D-ANAS) comprehensively on three typical public HSI datasets. The experimental results show that the HyT-NAS outperforms other state of the art DL based algorithms. Additionally, abundant ablation studies have been carried out to verify the effectiveness of the proposed  hybrid spatial-spectral search space and the grafted transformer. The results of the ablation study demonstrated that the HyT-NAS does find a local optimum architecture in the architecture search space. Compared with the pure CNN based HSI classification framework, the hybrid CNN-Transformer framework captures the global relationship between pixels. In future work, we will focus on designing a more efficient neural architecture search approach to automatically design a full transformer architecture for HSI classification.

\appendices



\ifCLASSOPTIONcaptionsoff
  \newpage
\fi




\bibliographystyle{IEEEtran}
\bibliography{reference}

\vspace{-0.8cm}
\begin{IEEEbiography}[{\includegraphics[width=1in,height=1.25in,clip,keepaspectratio]{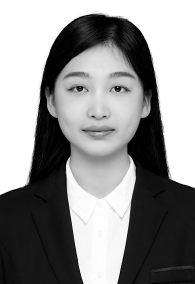}}]{Xizhe Xue} received the B.E. degree from Northwestern Polytechnical University, Xi’an, China in 2018. She is currently pursuing the Ph.D. degree in School of Computer Science, Northwestern Polytechnical University.

Her research interests include visual object tracking, HSI image processing, image segmentation techniques.

\end{IEEEbiography}

\vspace{-0.8cm}
\begin{IEEEbiography}[{\includegraphics[width=1in,height=1.25in,clip,keepaspectratio]{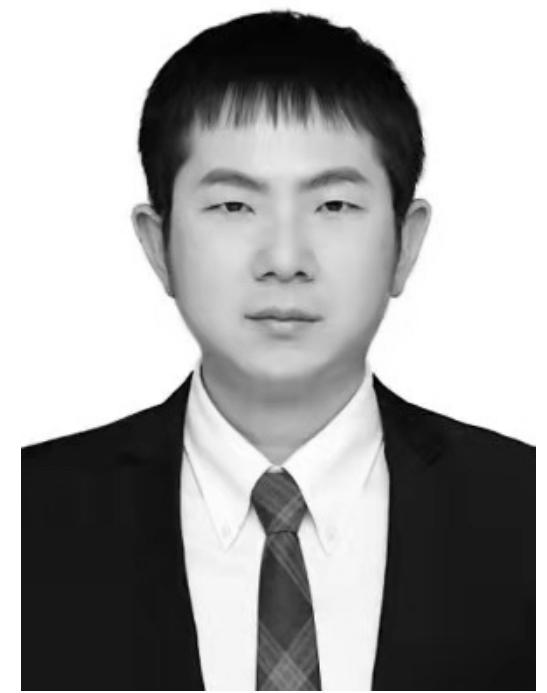}}]{Haokui Zhang}

received the PhD degree and the MS degree in computer application technology from Shannxi Provincial Key Laboratory of Speech and Image Information Processing in 2021 and 2016 respectively. He is currently work as a postdoctor in Harbin Institute of Technology, Shenzhen.

His research interests cover information retrieval, image restoration and hyperspectral image classification.
\end{IEEEbiography}

\begin{IEEEbiography}[{\includegraphics[width=1in,height=1.25in,clip,keepaspectratio]{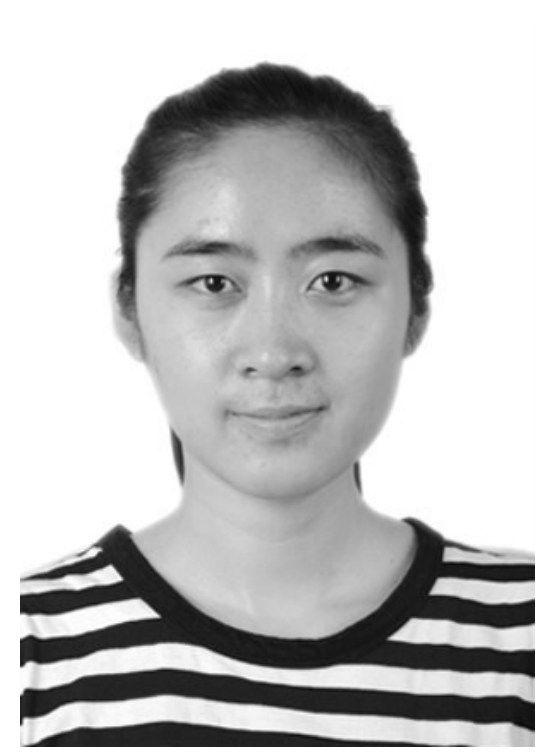}}]{Bei Fang}

received her Ph.D. degree in computer science and technology from the School of Computer Science at Northwestern Polytechnical University, Xi’an, China, in 2019. She is currently working as a Postdoctoral Research Associate with Key Laboratory of Modern Teaching Technology, Ministry of Education, Shaanxi Normal University, Xi’an, China.

Her research interests include computer vision, HSI image processing and deep learning techniques.

\end{IEEEbiography}

\begin{IEEEbiography}[{\includegraphics[width=1in,height=1.25in,clip,keepaspectratio]{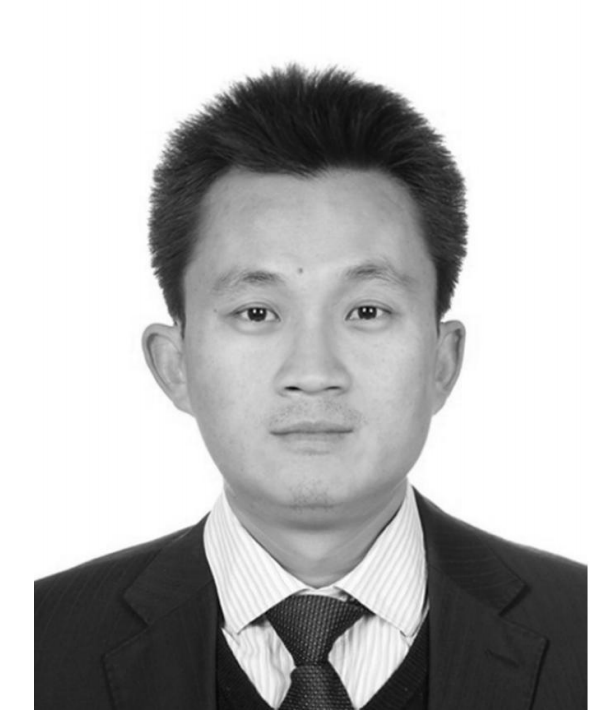}}]{Zongwen Bai}

received the MS degree from the Yan’an University, in 2008. He is currently pursuing the Ph.D degree with the School of Computer Science, Northwestern Polytechnical University, Xi’an,
China. He is an associate professor with the School of Physics and Electronic Information, Yan’an University, His research interests cover computer vision, nature language processing and deep learning.

His research interests include hyperspectral image super resolution, image fusion and deep learning.
\end{IEEEbiography}

\vspace{-0.8cm}

\begin{IEEEbiography}[{\includegraphics[width=1in,height=1.25in,clip,keepaspectratio]{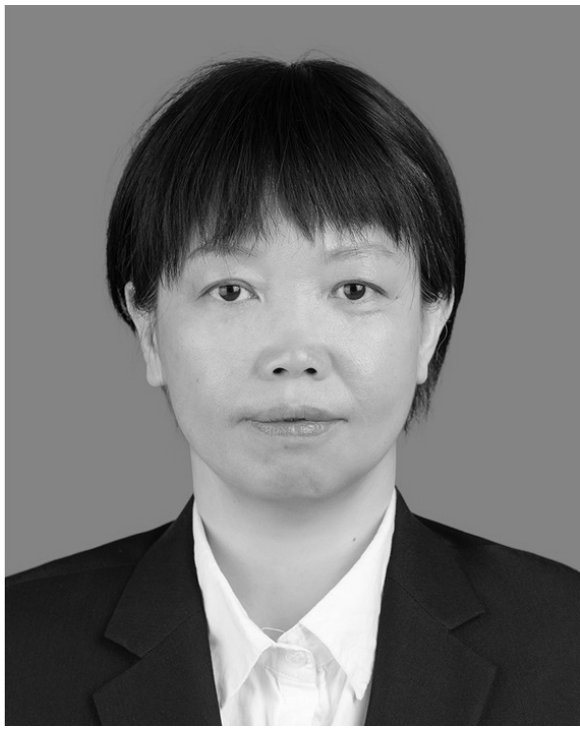}}]{Ying Li}

received PhD degree in electrical circuit and system from the National Key Laboratory of Radar Signal Processing, Xidian University in 2002.

Currently, she is a Professor with the School of Computer Science, Northwestern Polytechnical University. Her Interests cover image processing, computation intelligence and signal processing
\end{IEEEbiography}

\end{document}